\documentclass[sn-apa,iicol]{sn-jnl}% APA Reference Style 
\usepackage{graphicx}
\usepackage{multirow}
\usepackage{amsmath,amssymb,amsfonts,amsthm}
\usepackage{mathrsfs}
\usepackage[title]{appendix}
\usepackage{xcolor}
\usepackage{textcomp}
\usepackage{manyfoot}
\usepackage{booktabs}
\usepackage{algorithm}
\usepackage{algorithmicx}
\usepackage{algpseudocode}
\usepackage{listings}
\usepackage{subfigure}
\usepackage{times}
\usepackage{epsfig}
\usepackage[utf8]{inputenc}
\usepackage[T1]{fontenc}
\usepackage{url}
\usepackage{amsfonts}
\usepackage{nicefrac}
\usepackage{microtype}
\usepackage{siunitx}
\usepackage{comment}
\usepackage{indentfirst}
\usepackage{soul}
\usepackage{lipsum}
\usepackage{tabularx}
\usepackage{float}
\usepackage{xspace}
\usepackage{colortbl}

\newcommand{\new}[1]{\textcolor{black}{{#1}}}
\newcommand{\highlight}[1]{\textcolor{black}{{#1}}}

\makeatletter
\DeclareRobustCommand\onedot{\futurelet\@let@token\@onedot}
\def\@onedot{\ifx\@let@token.\else.\null\fi\xspace}

\def\eg{\emph{e.g}\onedot, } 
\def\ie{\emph{i.e}\onedot, }

 \def\vs{\emph{vs}\onedot}
 
\def\etal{\emph{et al}\onedot}

\makeatother

\theoremstyle{thmstyleone}%
\theoremstyle{thmstyletwo}%

\theoremstyle{thmstylethree}%

\begin{document}

\title[Article Title]{DO3D: Self-supervised Learning of Decomposed Object-aware 3D Motion and Depth from Monocular Videos}

\author[1]{\fnm{Xiuzhe} \sur{Wu}}\email{xzwu@eee.hku.hk}
\equalcont{These authors contributed equally to this work.}

\author[1]{\fnm{Xiaoyang} \sur{Lyu}}\email{shawlyu@connect.hku.hk}
\equalcont{These authors contributed equally to this work.}

\author[2]{\fnm{Qihao} \sur{Huang}}\email{qihao.huang@plus.ai}
\equalcont{These authors contributed equally to this work.}

\author[3]{\fnm{Yong} \sur{Liu}}\email{yongliu@iipc.zju.edu.cn}

\author[4]{\fnm{Yang} \sur{Wu}}\email{dylanywu@tencent.com}

\author[4]{\fnm{Ying} \sur{Shan}}\email{yingsshan@tencent.com}

\author[1]{\fnm{Xiaojuan} \sur{Qi}}\email{xjqi@eee.hku.hk}

% \affil[2]{\orgdiv{Department}, \orgname{Organization}, \orgaddress{\street{Street}, \city{City}, \postcode{10587}, \state{State}, \country{Country}}}

\affil[1]{\orgdiv{Department of Electrical and Electronic Engineering}, \orgname{The University of Hong Kong}, \orgaddress{\country{Hong Kong}}}

\affil[2]{\orgname{PlusAI corporation}, \orgaddress{\city{Shanghai}, \country{China}}}

\affil[3]{\orgdiv{The Institute of Cyber-Systems and Control}, \orgname{Zhejiang University}, \orgaddress{\city{Hangzhou}, \country{China}}}

\affil[4]{\orgname{Tencent corporation}, \orgaddress{\city{Shenzhen}, \country{China}}}


\abstract{Although considerable advancements have been attained in self-supervised depth estimation from monocular videos, most existing methods often treat all objects in a video as static entities, which however violates the dynamic nature of real-world scenes and fails to model the geometry and motion of moving objects. In this paper, we propose a self-supervised method to jointly learn 3D motion and depth from monocular videos. Our system contains a depth estimation module to predict depth, and a new decomposed object-wise 3D motion~(DO3D) estimation module to predict ego-motion and 3D object motion. Depth and motion networks work collaboratively to faithfully model the geometry and dynamics of real-world scenes, which, in turn, benefits both depth and 3D motion estimation. Their predictions are further combined to synthesize a novel video frame for self-supervised training. As a core component of our framework, DO3D is a new motion disentanglement module that learns to predict camera ego-motion and instance-aware 3D object motion separately. To alleviate the difficulties in estimating non-rigid 3D object motions, they are decomposed to object-wise 6-DoF global transformations and a pixel-wise local 3D motion deformation field. Qualitative and quantitative experiments are conducted on three benchmark datasets, including KITTI, Cityscapes, and VKITTI2, where our model delivers superior performance in all evaluated settings. For the depth estimation task, our model outperforms all compared research works in the high-resolution setting, attaining an absolute relative depth error~(abs rel) of \highlight{\textbf{0.099}} on the KITTI benchmark. Besides, our optical flow estimation results~(an overall EPE of \highlight{\textbf{7.09}} on KITTI) also surpass state-of-the-art methods and largely improve the estimation of dynamic regions, demonstrating the effectiveness of our motion model. Our code will be available.}

\keywords{3D motion decomposition, 3D motion estimation, depth estimation, 3D geometric consistency}



\maketitle

\section{Introduction}\label{sec1}
% What is the problem? Why is it important?
Estimating object-wise 3D motion and depth from monocular videos is a crucial yet challenging problem in outdoor scene understanding with many applications in autonomous driving vehicles and robots. Recently, supervised data-driven approaches with deep learning have shown promising results~\cite{li2023temporally,Teed2020RAFTRA,teed2020raft} for 3D motion and depth estimation. However, it is challenging to collect ground truth motion and depth data in large quantities, and models trained on limited training data also suffer from generalization issues~\cite{zhao2020towards} in {diverse} application scenarios. To address this limitation, self-supervised learning of depth~\cite{zhou2017unsupervised,godard2017unsupervised,mahjourian2018unsupervised,godard2019digging,liu2021self,zhao2022monovit,liu2023self} and motion~\cite{ren2017unsupervised,yin2018geonet,zou2018df,ranjan2019competitive} from a large amount of unlabeled monocular videos emerge as an alternative and scalable solution, which attracts a lot of research attention and demonstrate promising results.

% Why is the problem hard? What makes it challenging?
Despite recent advancements~\cite{godard2019digging,liu2021self,zhao2022monovit,liu2023self}, there are still several challenges in this area:
1) important scene structure information ({\eg} scale) is missing due to 3D-to-2D projections; 
2) 3D scene geometry ({\eg} depth), camera ego-motion, and object-wise motion are entangled together in monocular videos, making it hard to infer 3D motion and depth directly; and
3) 3D non-rigid deformation patterns ({\eg} pedestrian motion) are diverse and complicated, {causing it difficult} to learn in a self-supervised manner.

\begin{figure}
  \centering
    \begin{tabular}{cc}
	
    \hspace{-3mm}
    \begin{minipage}[b]{0.48\linewidth}
	    \centering
	    {\includegraphics[width=\linewidth]{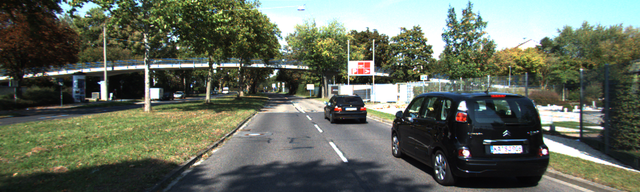}}
	\end{minipage} &
	\hspace{-5mm}
	\begin{minipage}[b]{0.48\linewidth}
	    \centering
	    {\includegraphics[width=\linewidth]{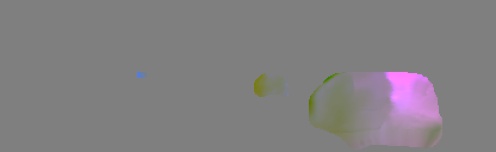}}
	\end{minipage} \\
	\footnotesize Source & \footnotesize GeoNet \\
	\hspace{-3mm}
	\begin{minipage}[b]{0.48\linewidth}
	    \centering
	    {\includegraphics[width=\linewidth]{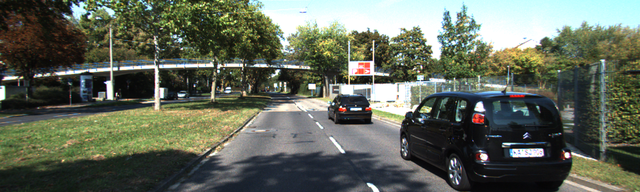}}
	\end{minipage} &
	\hspace{-5mm}
	\begin{minipage}[b]{0.48\linewidth}
	    \centering
	    {\includegraphics[width=\linewidth]{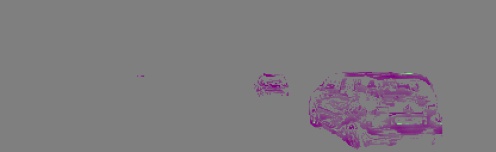}}
	\end{minipage} \\
	\footnotesize Target & \footnotesize {Li \etal} \\
	\hspace{-3mm}
	\begin{minipage}[b]{0.48\linewidth}
	    \centering
	    {\includegraphics[width=\linewidth]{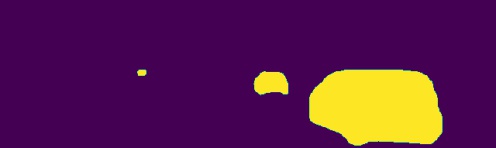}}
	\end{minipage} & 
	\hspace{-5mm}
	\begin{minipage}[b]{0.48\linewidth}
	    \centering
	    {\includegraphics[width=\linewidth]{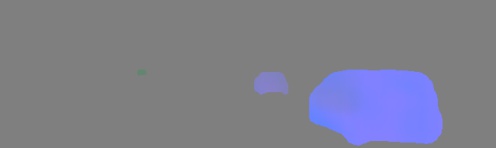}}
	\end{minipage} \\
	\footnotesize Mask & \footnotesize Ours \\
    \end{tabular}
    \caption{{Visualization of 3D motion fields generated by} GeoNet~\cite{yin2018geonet}, {Li \etal}~\cite{li2020unsupervised}, and {our model}. Our decomposed motion model predicts a 3D motion {field more consistent} than {the} others {obtained} by optical flow or direct methods. R, G, B color maps correspond to motion in x, y, z directions, respectively. 
    }
    
	\label{fig:3d_motion}
\end{figure}

\begin{figure*}[t]
    \centering
    \includegraphics[width=1.0\linewidth]{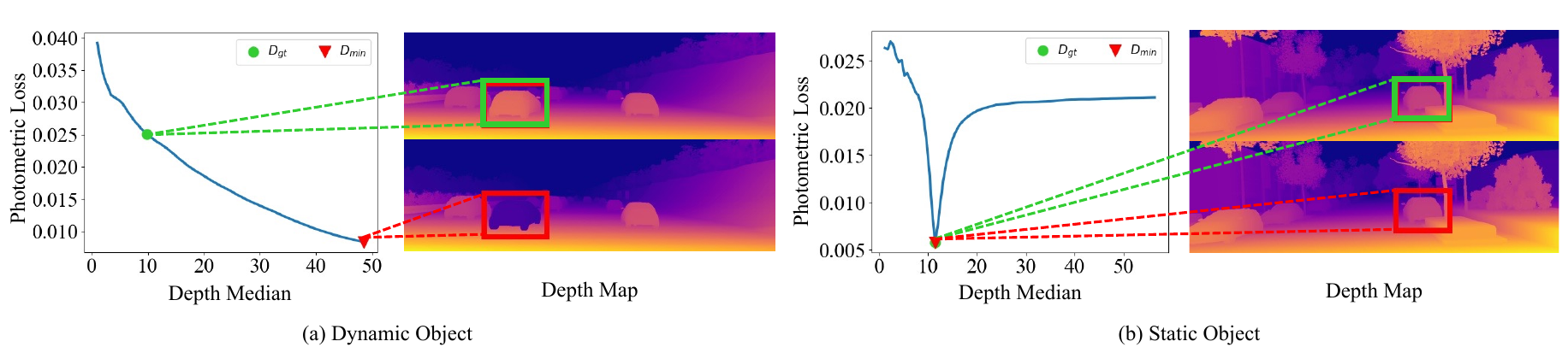}
    % \vspace{-0.7cm}
    \caption{Depth-loss curves of the dynamic object~(a) and the static object~(b). $D_{gt}$ and $D_{min}$ are the points {indicating the} depth ground truth and {the} minimum photometric loss, respectively.}
    \label{fig:depth_loss_curve}
\end{figure*}

To this end, a large body of recent research \cite{godard2019digging,lyu2020hr,shu2020feature,guizilini2020semantically,jung2021fine,liu2021self,zhao2022monovit,liu2023self} simplifies the real-world model, assuming that the scene is static and thus avoids the difficulties in jointly modeling motion and scene structures. However, such assumptions violate the dynamic nature of real-world scenes and will inevitably lead to errors in depth and motion estimation of dynamic objects; {see Figures \ref{fig:depth_loss_curve} and \ref{fig:blackhole}}.
For instance, the estimated depth for dynamic objects often appears to be black~({\ie} black hole issue), indicating an infinity value. 
This is caused by the fact that the static-scene formulation breaks the photo-consistency assumption of dynamic scenes and misleads the self-supervised depth estimation networks to produce inaccurate predictions.
To address the above issue, the following works \cite{cao2019learning,lee2019instance,li2020unsupervised, yin2018geonet,ranjan2019competitive,xu2021moving} attempt to incorporate motion into the self-supervised learning framework.
{In this line of research, dedicated modules are developed to predict object-wise motion~\cite{cao2019learning,lee2019instance}, pixel-wise motion~\cite{li2020unsupervised,yin2018geonet,ranjan2019competitive} or patch-wise motion~\cite{xu2021moving}. However, the above approaches either entangle individual object-/patch-wise motion with camera motion~\cite{cao2019learning, xu2021moving}, ignoring motion consistency within an object area
% \textcolor{red}{this point is a bit difficult to understand, why ignores motion consistency within an object}, 
{({\eg} pixels within an object area often follow similar motion patterns)~\cite{li2020unsupervised},}
or just model motion in 2D space using optical flow~\cite{yin2018geonet,ranjan2019competitive}, overlooking the fact that motion happens in 3D space and makes the prediction challenging.}
Therefore, motion estimation still suffers from an unsatisfactory performance, as shown in Figure \ref{fig:3d_motion}.

% What does our paper contribute? What is the key idea? What is the magic trick? What is the new insight or technique that enables us to advance the frontier?
To shed light on a better understanding of the self-supervised depth estimation in dynamic scenes, we first conduct theoretical and empirical analyses of the widely adopted self-supervised learning framework~\cite{zhou2017unsupervised,godard2017unsupervised,mahjourian2018unsupervised,godard2019digging,lyu2020hr,shu2020feature,guizilini2020semantically,liu2021self,zhao2022monovit,liu2023self}. We mathematically prove that such issues cannot be avoided in existing self-supervised learning formulations and will become more destructive in highly dynamic scenes: details are unfolded in Section~\ref{sec: analysis}.
To this end, we present a new formulation based on the underlying 3D geometry and motion model: \textit{a scene is a composition of objects and background in 3D space; the combination of ego-motion and object motion determines the evolution of the dynamic 3D scene with respect to an observer, a video is a projection of the composed 3D dynamic scene onto the camera image plane determined by camera pose, i.e., ego-motion.} 
To mimic such a process, we propose to disentangle geometry, 3D object motion, and camera ego-motion, and then compose them together to synthesize a new 3D scene. The new 3D scene is finally projected to the camera plane to generate a video frame for constructing the self-supervised learning objective. 

Concretely, the disentanglement can be achieved by separately predicting depth for 3D scene geometry, camera ego-motion, and object motion. 
For depth estimation, it is noticed that the performance of depth estimation is highly related to the representative capabilities of the model, which requires global information for resolving ambiguities and local information for details. Inspired by the recent success of vision transformers in dense prediction tasks ~\cite{liu2021swin,liu2022swin,xie2021segformer}, we design a transform-based encoder network~\cite{xie2021segformer} to extract global features and a CNN-based decoder to facilitate detailed predictions, serving as a strong baseline backbone.
For camera ego-motion and object motion prediction, we find it difficult to directly predict object motion using neural networks due to the entanglement and diverse nature of real-world 3D motions. We thus propose a new 3D motion disentanglement module, namely decomposed object-aware 3D motion (DO3D) prediction, for camera ego-motion and object motion prediction. This serves as an effective geometric regularization for motion disentanglement and prediction. 

In detail, DO3D first predicts camera ego-motion, which aligns the coordinate system of nearby video frames. 
Then, to model complicated 3D object motion, we formulate the 3D object-wise motion as the composition of a global object-wise 6-DoF rigid transformation and a 3D pixel-wise motion deformation. The pixel-wise deformation refines the results and produces high-quality motion for non-rigid objects such as cyclists and pedestrians.
Our formulation is inspired by the observation that the motion of many objects in outdoor scenes is globally rigid, {\eg} cars with small local variations or pedestrian movements, as shown in {Figure}~\ref{fig:s123_comparison}.
Thanks to our formulation, the proposed approach can produce high-quality 3D object-wise motion {fields} and depth {maps as} shown in {Figure}~\ref{fig:3d_motion}, Figure~\ref{fig:depth_visualization_cs}, and Figure~\ref{fig:blackhole}. 

We evaluate our model on three autonomous driving datasets, including KITTI~\cite{geiger2012we}, Cityscapes~\cite{cordts2016cityscapes}, and VKITTI2~\cite{cabon2020vkitti2,gaidon2016virtual}.
Compared with KITTI, Cityscapes and VKITTI2 datasets contain more dynamic objects, {and VKITTI2 provides a more precise ground truth~(\ie depth and optical flow).}
{We attain state-of-the-art depth estimation performance on KITTI in both high- and low-resolution settings~(abs rel of \highlight{0.116} in low-resolution setting and abs rel of \highlight{0.099} in high-resolution setting).
Besides, thanks to our motion estimation module, we obtain more performance gains in regions that contain dynamic objects where the RMSE metric decreases more in foreground regions from \highlight{4.814} to \highlight{4.737}.}
For motion estimation, our method outperforms existing approaches with an overall EPE of \highlight{7.09} in optical flow estimation and a bad pixel percentage of \highlight{54.68\%} in scene flow estimation.

Our major contributions are summarized below:
% What do the experiments say?
\begin{itemize}
	\item [1)] 
        We present an analysis of existing formulations for self-supervised depth estimation, demonstrate the root causes of their failure modes, and propose a new self-supervised framework to jointly learn decomposed object-wise 3D motion and dense scene depth from monocular videos, attempting to model the underlying geometric model.
    
    \item [2)]
        We contribute a better baseline backbone network with a hybrid transformer and CNN architecture for depth estimation. The transformer component effectively {exploits spatial correlations between regions within a frame and captures global information, which achieves significantly better performance than previous works.}
	\item [3)]
        We develop a new 3D motion estimation method with disentanglement, namely DO3D, to predict camera ego-motion, object-wise rigid motion, and non-rigid deformation, exploiting real-world motion constraints to regularize motion predictions. DO3D is generic and can be incorporated into existing state-of-the-art self-supervised depth estimation methods for better motion and depth estimation.
	
	\item [4)]
        We evaluate our model on depth, optical flow, and scene flow estimation tasks.
        Quantitative and qualitative results on three driving datasets show the superiority of our approach, especially in highly dynamic scenarios. 
\end{itemize}

The paper is organized as follows. Section~\ref{sec:related_work} summarizes recent related works in depth and motion estimation tasks. Then, section~\ref{sec:background} presents rigorous analyses of existing formulations and unresolved challenges. 
Further, we elaborate on our proposed framework in Section~\ref{sec:model}, consisting of a hybrid transformer and CNN architecture for depth estimation and a 3D motion estimation module. Experimental results and ablation studies are shown in Section~\ref{sec:experiments}. Finally, we analyze our limitations in Section~\ref{sec: limitation} and draw a conclusion in Section ~\ref{sec:conclusion}.

\section{Related Work}\label{sec:related_work}

In the following, we start by reviewing {supervised depth estimation methods}. Then, we review related works in self-supervised monocular depth estimation. Finally, we introduce existing works for motion estimation of dynamic objects. 

\noindent{\textbf{Supervised Depth Estimation Methods.}}
Monocular depth estimation from a single image is an ill-posed problem, suffering from scale ambiguities.
Many works~\cite{eigen2015predicting,laina2016deeper} leverage supervised learning methods to let the neural network learn depth information from training data. 
Eigen {\etal} \cite{eigen2015predicting} formulate depth estimation from a single image as a regression task and develop a fully convolutional coarse-to-fine network to predict pixel-wise depth values. Later, Laina {\etal} \cite{laina2016deeper} develop a deep residual network to improve depth estimation and achieve higher accuracy. However, supervised approaches rely on large-scale data with ground truth depth value and suffer from poor generalization. Therefore, the community has started to explore self-supervised learning for depth estimation, which doesn't require labeled datasets.

\noindent {\textbf{Self-supervised monocular depth estimation.}}
The goal of self-supervised monocular depth estimation is to predict a dense depth map for each single input image in a self-supervised manner. The key idea is to use predicted scene depth and camera ego-motion to reconstruct a target frame given a source frame observation using differentiable warping. Then, the model is trained by minimizing the reconstruction error. The first self-supervised depth estimation framework is proposed in~\cite{garg2016unsupervised} for stereoscopic videos and extended to handle monocular videos in~(SfMLearner)~\cite{zhou2017unsupervised}. Monodepth2~\cite{godard2019digging} further improves it by handling occlusion, filtering relatively static scenes, and introducing additional constraints, such as multi-scale appearance matching loss, to further improve the estimation accuracy. Bian \etal~\cite{bian2019unsupervised} propose a depth consistency loss to avoid scale inconsistency among neighboring frames. Mahjourian \etal~\cite{mahjourian2018unsupervised} put forth a geometric loss based on a classical rigid registration method-- Iterative Closest Point~(ICP). Semantic information is also exploited as constraints~\cite{guizilini2020semantically, jung2021fine}. Shu \etal~\cite{shu2020feature} present a feature-metric loss to measure the reconstruction error in deep feature space. Another way to boost model performance is to improve the structure of backbone networks. Some works develop better encoders that change~(usually ResNet18) to a stronger one~(\eg ResNet50~\cite{jung2021fine}, ResNet101~\cite{johnston2020self} or PackNet~\cite{guizilini20203d, guizilini2020semantically}) while others enhance the complexity of  decoder to capture fine details~\cite{lyu2020hr}. Furthermore, since neighboring frames are often available at inference time, a few recent works use multiple frames for depth estimation to leverage temporal information. For instance, \cite{wang2019recurrent,patil2020don,cs2018depthnet,zhang2019exploiting} process frame sequences by introducing the recurrent neural layer~(\eg LSTM) to model the long-term dependency. Watson \etal \cite{watson2021temporal} build a cost volume inspired by multi-view stereo methods. This paper focuses on depth estimation utilizing a single frame. We build a powerful transformer-based depth encoder and an efficient convolution-based depth decoder to improve depth estimation from a single frame.

\noindent \textbf{Motion estimation for dynamic objects.}
One major drawback of the previous self-supervised depth estimation method is that they are incapable of handling moving objects.
To address the issue, existing attempts~\cite{godard2019digging, zhou2017unsupervised, yin2018geonet, ranjan2019competitive} can be coarsely categorized into the following lines.
One line of research tries to remove the dynamic areas when calculating the self-supervised loss. For instance, the auto-masking scheme~(\ie ignoring moving objects at the same speed) ~\cite{godard2019digging} and semantic mask prediction~\cite{zhou2017unsupervised} are adopted to ignore dynamic objects in network optimization. 
However, by simply discarding these regions, the model can not effectively learn to model dynamic objects and object motion, leading to failures in highly dynamic scenes on depth and motion estimation. 

Another line is to explicitly model motion by learning 2D optical flows, which compensate for the influence of dynamic objects. Yin~\etal~\cite{yin2018geonet} predict a residual flow map to revise 2D correspondences computed using depth and camera poses. Ranjan \etal~\cite{ranjan2019competitive} incorporate an extra flow estimation network to estimate 2D correspondences in moving regions and train the whole system in a competitive and collaborative manner. However, this formulation overlooks the underlying geometry because the natural motion happens in 3D. As a result, the performance suffers from limitations in optical flow estimation, and it is still difficult to infer the explicit 3D motion pattern from the residual flow, as shown in Figure~\ref{fig:3d_motion}.

The third line of research most related to our work is to model the 3D motion of dynamic objects. A few recent methods~\cite{casser2019struct2depth, cao2019learning, li2020unsupervised, lee2019instance, xu2021moving} attempt to estimate motion using self-supervised learning by disentangling the camera and object motion. Some works predict pixel-wise scene flow maps to model object motion~\cite{cao2019learning, li2020unsupervised}. Cao~\etal~\cite{cao2019learning} propose a self-supervised framework with given 2D bounding boxes to learn 3D object motion from stereo videos. Li~\etal ~\cite{li2020unsupervised} directly predict a pixel-wise 3D motion map for a frame. This scheme lacks object-level regularization and has too much degree of freedom to learn, making it not effective to incorporate priors, \eg most object motion is piece-wise rigid, into the model. Meanwhile, the other stream of methods~\cite{casser2019struct2depth, lee2019instance} follows the rigid motion assumption. They estimate a 6 Degree-of-Freedom parameter to represent object motion. However, they directly output the overall motions in the object area, which means the output motions are still entangled with camera poses. Besides, they all lack consideration of non-rigid motions. Most recently, moving SLAM~\cite{xu2021moving} predicts poses for each locally rigid region and further uses a learned foreground segmentation network for view synthesis. Such a framework can not reflect precise object-wise motion prediction, especially in complicated driving scenes.

In contrast to existing methods, our method is based on monocular frames. We attempt to resolve this issue by modeling the underlying 3D geometric model and disentangling geometry, camera motion, and object motion. Besides, We dig deeper into object motion modeling: previous works either predict a non-rigid motion map without any restriction for all the objects or focus on motion estimation for rigid objects. Differently, our decomposed motion estimation module is designed to model different kinds of object motion in one framework, including both rigid motions and pixel-wise deformations as motion compensation. Furthermore, our proposed DO3D module can serve as a plug-and-play component and be incorporated into existing methods~\cite{godard2019digging,lyu2020hr,shu2020feature} for better self-supervised learning of depth and motion.

\section{Background and Challenges}\label{sec:background}
We present background knowledge of the self-supervised monocular depth estimation task in Section~\ref{sec:pre}, conduct empirical and theoretical analysis in Section~\ref{sec:empirical} \& Section~\ref{sec: analysis}, and highlight unresolved challenges in Section~\ref{sec:unresolved}.

\subsection{Preliminaries}\label{sec:pre}
\begin{figure}
    \centering
    \includegraphics[width=1\linewidth]{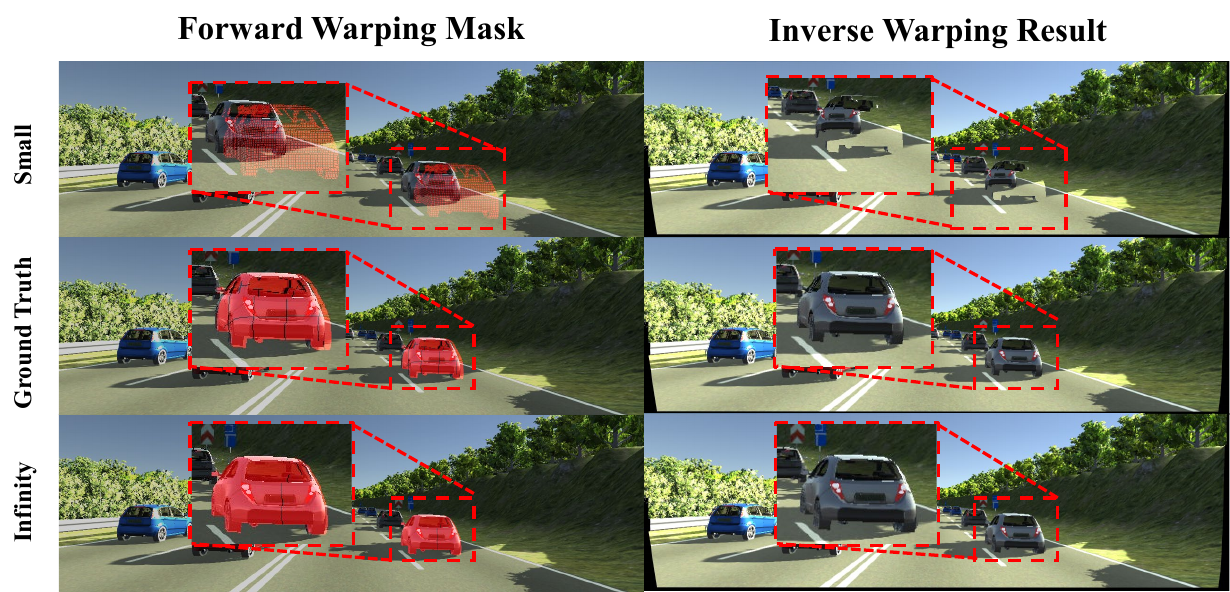}
    \caption{
    Warping process visualization. We visualize two important results of inverse warping at different depths. We set the smallest depth to $0.1\times d_{gt}$ and infinity to $5 \times d_{gt}$. The forward warping mask is computed by equation \eqref{eq:inverse_warpping} and the red mask means projected coordinates $\mathrm{u}_s$ and $\mathrm{v}_s$.}
    \label{fig:dynamic_warping}
\end{figure}

The framework of self-supervised depth estimation~\cite{zhou2017unsupervised,godard2017unsupervised,mahjourian2018unsupervised,godard2019digging,lyu2020hr,shu2020feature,guizilini2020semantically} is mainly composed of depth prediction network (i.e., DepthNet) and camera pose estimation network (i.e., PoseNet). It aims to utilize the geometric relationships obtained from video sequences to reconstruct 3D scene geometry. If the source video frame is denoted as $\mathrm{I}_s$, $s\in \{t-1, t+1\}$ and the target frame as $\mathrm{I}_t$, their geometric relationship can be represented as 
\begin{equation}
	\mathrm{d}_s\mathrm{p}_s = \mathrm{KT}_{t\rightarrow s}(\mathrm{d}_t \mathrm{K}^{-1} \mathrm{p}_t),
	\label{eq:inverse_warpping} 
\end{equation}
where $\mathrm{p}_t$ and $\mathrm{p}_s$ are defined as 2D homogeneous pixel grid coordinates in $\mathrm{I}_t$ and $\mathrm{I}_s$, respectively, $\mathrm{d}_t$ and $\mathrm{d}_s$ are the corresponding depth values of $\mathrm{p}_t$ and $\mathrm{p}_s$ predicted by DepthNet, $\mathrm{T}_{t\rightarrow s}$ is the camera extrinsic matrix predicted by PoseNet, and $K$ is the camera intrinsic matrix. 
{Given the correspondence between a source pixel $\mathrm{p}_s$ and a target pixel $\mathrm{p}_t$ as described in Equation~\eqref{eq:inverse_warpping},} it is easy to map source pixels $\mathrm{I}_s$ to the target frame $\mathrm{I}_t$ where we use the bilinear interpolation method to reconstruct the source frame and denote the reconstructed source frame as $\mathrm{\hat{I}}_{t}$. 
As in~\cite{godard2019digging}, DepthNet and PoseNet are trained to minimize the photometric loss, as shown in Equation~\eqref{eqn:photo_metric_loss},
\begin{equation}
	\small
	\mathrm{\mathcal{L}}_{\mathrm{ph}}(\mathrm{\hat{I}}_t, \mathrm{I}_t) = \frac{\mathrm{\alpha}}{2}(1-\text{SSIM}(\mathrm{\hat{I}}_{t}, \mathrm{I}_{t})) + (1-\mathrm{\alpha}) \Vert \mathrm{\hat{I}}_{t} - \mathrm{I}_{t} \Vert_1, 
	\label{eqn:photo_metric_loss} 
\end{equation}
\noindent where $\alpha$ is a hyper-parameter balancing the SSIM~\cite{wang2004image} term and $l_1$ pixel-wise differences. In the following, we present an empirical and theoretical analysis of the above formulation for self-supervised depth estimation.
    
\subsection{Empirical Analysis}~\label{sec:empirical}
Although the above photometric loss has been widely adopted in existing works~\cite{garg2016unsupervised, zhou2017unsupervised, godard2019digging}, it assumes that all pixels are static and thus violate the dynamic nature of real-world scenes. In this section, we conduct experimental studies on its negative impacts on the depth estimation of dynamic objects.

In our initial experiments, we investigate whether minimizing the photometric loss, as defined in Equation~\eqref{eqn:photo_metric_loss}, will effectively encourage the model to generate accurate depth estimations. The photometric loss is a commonly employed loss function in self-supervised depth estimation tasks. It is well noted that the preliminary experiment is conducted on a pre-trained model with fixed weights. To produce this curve, we uniformly sample depth values within a valid range, utilize this depth map to warp the source image into the target camera plane, and subsequently calculate the corresponding photometric loss between the warped image and the target image. Figure~\ref{fig:depth_loss_curve} depicts two curves corresponding to different scenes: one featuring static objects and the other with dynamic objects. Within these curves, we have identified two critical points. The first is the ground-truth depth point, represented by green points. The second is the point with the minimum photometric loss, indicated by a red triangle. Notably, even when using the ground-truth depth for dynamic objects, the photometric loss fails to reach its minimum value, as shown in Figure~\ref{fig:depth_loss_curve} (a). In contrast, the loss value gradually decreases as the depth value approaches infinity. However, this trend does not occur in static areas, where the ground-truth depth leads to the minimum photometric loss, as demonstrated in Figure~\ref{fig:depth_loss_curve} (b). These figures illustrate that the depth corresponding to the minimum loss does not coincide with the true depth of dynamic objects. As a result, within the current framework, optimizing photometric loss directly without incorporating motion modeling for dynamic objects would cause the model to acquire incorrect knowledge, leading to inaccurate depth estimation.

Further, the phenomenon that assigning an infinity depth value to dynamic objects will give rise to a small photometric loss is further verified by our warping experiment. Specifically, we select a dynamic scene from VKITTI2 and visualize the warped images of moving objects at different depth values. Visualization results are shown in Figure~\ref{fig:dynamic_warping}. The first row shows that a smaller depth value will make the projected coordinates far from the dynamic objects and lead to erroneous warping results. However, if we assign an infinite depth to the pixels of corresponding dynamic objects, the warped images become better, leading to a much smaller photometric loss. Although there are several recent efforts~\cite{godard2019digging, luo2019every, casser2019struct2depth, guizilini20203d} attempting to solve it, a rigorous mathematical analysis of why this happens is still lacking. Instead, we provide an analysis of this issue in Section~\ref{sec: analysis}, which motivates our proposed approach.

\begin{figure*}
    \centering
    \includegraphics[width=1\linewidth]{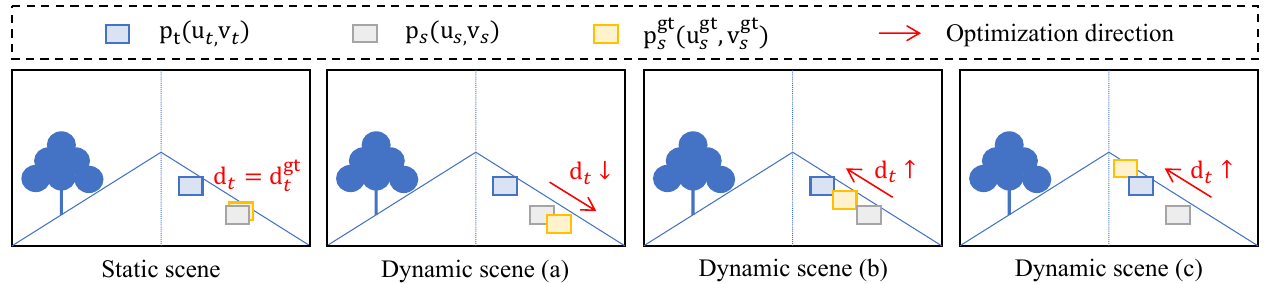}
    % \vspace{-0.7cm}
    \caption{{Visualization of four motion statuses. For simplification, we assume that the ego-car is moving forward and only analyze the situations where the target pixels are located in the right part of the image. 
    Let $p_t(\mathrm{u}_t,\mathrm{v}_t)$ be the pixel in the target frame and $\mathrm{p}_s$ be the projected pixel in the source frame computed by predicted depth $\mathrm{d}_t$ and camera pose according to {Equation \eqref{eq:inverse_warpping}}. 
    $\mathrm{p}_s^\mathrm{gt}$ denotes the observed source pixel under different motion patterns of the object.
    For static scenes that follow the self-supervised depth estimation model, $\mathrm{p}_s^\mathrm{gt}$ and $\mathrm{p}_s$ will be the same point.
    In dynamic scene (a) -- (c), $\mathrm{p}_s$  will not be at the same location as $\mathrm{p}_s^\mathrm{gt}$. However, the photometric loss will encourage the model to produce the estimated depth $\mathrm{d}_t$ that enforces $\mathrm{p}_s$ to approach $\mathrm{p}_s^\mathrm{gt}$, which will mislead depth estimation optimization.}}
    \label{fig:black_hole_theory}
\end{figure*}

\subsection{Mathematical Analysis}
\label{sec: analysis}
In this section, we analyze the mathematical formulation of the self-supervised depth estimation objective and its impacts on depth estimation in real-world dynamic scenes.
We will first describe the self-supervised depth estimation model under the static scene assumption. Then, we will analyze the geometric model of real-world dynamic scenes. 
Finally, we obtain the mathematical relationship between the true depth $\mathrm{d}_t^{\mathrm{gt}}$ and the estimated depth $\mathrm{d}_t$ and conduct an analysis.
In the following, given a target pixel $(\mathrm{u}_t, \mathrm{v}_t)$,  we use $(\mathrm{u}_s, \mathrm{v}_s)$ to represent the source pixel location obtained according to estimated depth $
\mathrm{d}_t$ from DepthNet, and $(\mathrm{u}_s^\mathrm{gt}, \mathrm{v}_s^\mathrm{gt})$ to denote the corresponding observed coordinate obtained according to true depth $\mathrm{d}_t^\mathrm{gt}$ and real-world dynamic scene model.

\noindent\textbf{Self-supervised depth estimation model under the static scene assumption.} 
Equation~\eqref{eq:inverse_warpping} expresses the relationship between two distinct image planes, assuming a static scene. To simplify the representation, we define $\mathrm{u}_x$ and $\mathrm{v}_y$, given the 2D location of the target pixel $\mathrm{p}_t (\mathrm{u}_t, \mathrm{v}_t)$:
\begin{equation}
    \begin{aligned}
    \mathrm{u}_x &= \frac{1}{\mathrm{f}_x}(\mathrm{u}_t - \mathrm{c}_x), \\
    \mathrm{v}_y &= \frac{1}{\mathrm{f}_y}(\mathrm{v}_t - \mathrm{c}_y),
    \end{aligned}
    \label{eq:u_x_v_y}
\end{equation}
where $(\mathrm{c}_x, \mathrm{c}_y)$ denotes {the principal point}. Then, {by expanding Equation~\eqref{eq:inverse_warpping} and using Equation~\eqref{eq:u_x_v_y} to simplify the notation,}  we can obtain the new projected location  $\mathrm{p}_s = (\mathrm{u}_s, \mathrm{v}_s)$ on the source camera plane as
\begin{equation}
    \begin{aligned}
    \mathrm{u}_s &=\mathrm{f}_x \cdot \frac{\mathrm{r}_{11}\mathrm{u}_x \mathrm{d}_t + \mathrm{r}_{12}\mathrm{v}_y \mathrm{d}_t + \mathrm{r}_{13}\mathrm{d}_t + \mathrm{t}_1}{\mathrm{r}_{31}\mathrm{u}_x \mathrm{d}_t + \mathrm{r}_{32}\mathrm{v}_y \mathrm{d}_t + \mathrm{r}_{33}\mathrm{d}_t + \mathrm{t}_3} + \mathrm{c}_x  \\
    \mathrm{v}_s &=\mathrm{f}_y \cdot \frac{\mathrm{r}_{21}\mathrm{u}_x \mathrm{d}_t + \mathrm{r}_{22}\mathrm{v}_y \mathrm{d}_t + \mathrm{r}_{23}\mathrm{d}_t + \mathrm{t}_1}{\mathrm{r}_{31}\mathrm{u}_x \mathrm{d}_t + \mathrm{r}_{32}\mathrm{v}_y \mathrm{d}_t + \mathrm{r}_{33}\mathrm{d}_t + \mathrm{t}_3} + \mathrm{c}_y. 
    % d' &=r_{31}u_x + r_{32}v_y + r_{33}\mathrm{d}_t + \mathrm{t}_3.
    \end{aligned}
    \label{eq:u'}
\end{equation}
where $\mathrm{r}_{ij}$ and $\mathrm{t}_i$ are the elements of relative pose $\mathrm{T}_{t\rightarrow s}([R|t])$ of the new camera viewpoint with respect to the original one, and $\mathrm{d}_t$ is the depth prediction result from DepthNet.

Considering that in most autonomous driving scenarios, the ego-motion is dominated by the  straight movement of the ego car ({\ie} translation component only), we can approximate the extrinsic matrix as
\begin{equation}
    \mathrm{R} \approx \left[
    	\begin{matrix}
    	\mathrm{r}_{11} = 1 & \mathrm{r}_{12} = 0 & \mathrm{r}_{13} = 0 \\
    	\mathrm{r}_{12} = 0 & \mathrm{r}_{22} = 1 & \mathrm{r}_{23} = 0 \\
    	\mathrm{r}_{13} = 0 & \mathrm{r}_{32} = 0 & \mathrm{r}_{33} = 1
    	\end{matrix}
      \right],\\
    \mathrm{t} = \left[
    		\begin{array}{c}
    			\mathrm{t}_1 \\
    			\mathrm{t}_2 \\
    			\mathrm{t}_3
    		\end{array}
    	\right].
	\label{eq:simplified_rt}
\end{equation}
Then, taking $\mathrm{u}_s$ as an example, by plugging Equation~\eqref{eq:simplified_rt} into Equation~\eqref{eq:u'}, we can obtain $\mathrm{u}_s$ as
\begin{align}
    \mathrm{u}_s = \frac{\mathrm{d}_t(\mathrm{u}_t-\mathrm{c}_x) + \mathrm{f}_x \cdot \mathrm{t}_1}{\mathrm{d}_t + \mathrm{t}_3} + \mathrm{c}_x.
    \label{eq:simplified_u_s}
\end{align}
Equation~\eqref{eq:simplified_u_s} shows the relationship between the depth value $\mathrm{d}_t$ and pixel abscissa $\mathrm{u}_s$ given a known ego-motion $\mathrm{T}_{t\rightarrow s}([R|t])$. 
Without loss of generality, we assume the ego camera is moving forward with only motion along the z-axis, {\ie}  $\mathrm{t}_1 = 0, \mathrm{t}_2 = 0, \mathrm{t}_3 < 0$,
Equation~\eqref{eq:simplified_u_s} becomes
\begin{align}
    \label{eq:simplified_u_s_new}
    \mathrm{u}_s &= \frac{\mathrm{d}_t\cdot(\mathrm{u}_t-\mathrm{c}_x)}{\mathrm{d}_t + \mathrm{t}_3} + \mathrm{c}_x,
\end{align}
and the partial derivative of $\mathrm{u}_s$ with respect to $\mathrm{d}_t$ is
\begin{equation}\label{eq: derivate u2d}
    \frac{\mathrm{\partial}{\mathrm{u}_s}}{\mathrm{\partial}{\mathrm{d}_t}} = \frac{(\mathrm{u}_t-\mathrm{c}_x)\mathrm{t}_3}{(\mathrm{d}_t + \mathrm{t}_3)^2}.
\end{equation}
The sign of Equation~\eqref{eq: derivate u2d} is
\begin{equation}\label{eq:signal_function}
\mathrm{sign}(\frac{\mathrm{\partial}{\mathrm{u}_s}}{\mathrm{\partial}{\mathrm{d}_t}}) = 
\begin{cases}
\mathrm{negative}, & \text{if $\mathrm{u}_t > \mathrm{c}_x$} \\
\mathrm{positive}, & \text{if $\mathrm{u}_t < \mathrm{c}_x$}.
\end{cases}
\end{equation}
It indicates that if a pixel is on the right half side of the image ({\ie} $\mathrm{u}_t > \mathrm{c}_x$): 1) as $\mathrm{d}_t$ becomes smaller, the projected abscissa ${\mathrm{u}_s}$ will become larger, and the projected point will be closer to the right border of the image; and 2) as $\mathrm{d}_t$ turns larger and approaches positive infinity, the projected abscissa ${\mathrm{u}_s}$ will become smaller and approach the original location $\mathrm{u}_t$ as
\begin{equation}
    \lim \limits_{\mathrm{d}_t\rightarrow \infty}\mathrm{u}_s = \mathrm{u}_t,
    \label{eq:limit_u}
\end{equation}
In summary, if a point $\mathrm{u}_t$ is on the right-hand side of the image ($\mathrm{u}_t > \mathrm{c}_x$), the projected point $\mathrm{u}_s$ will be on the right of $\mathrm{u}_t$ ({\ie} $\mathrm{u}_s \ge \mathrm{u}_t$). Conversely, if $u < c_x$,  $\mathrm{u}_s \le u$. 
The above analysis can also be generalized to the vertical coordinate $\mathrm{v}_t$.
Given the above facts, we analyze the impact of object motion on self-supervised depth estimation. 

\noindent\textbf{Real-world dynamic scene model.} 
Suppose that the 3D location of the target pixel is $(x_t, y_t, \mathrm{d}_t^{\mathrm{gt}})$, and the object moves along the $z$-axis for $\Delta{t}_3$,  the true location of the pixel at the source frame time-step is $(x_t, y_t, \mathrm{d}_t^{\mathrm{gt}}+\Delta{t}_3 + t_3)$. According to the pin-hole camera model, the following can be derived
\begin{equation}
    \begin{aligned}
    \mathrm{u}_t &= \frac{{x}_t\mathrm{f}_x}{\mathrm{d}_t^\mathrm{gt}} + \mathrm{c}_x, \\
    \mathrm{u}_s^{\mathrm{gt}} &= \frac{x_t \mathrm{f}_x}{\mathrm{d}_t^\mathrm{gt}+\Delta \mathrm{t}_3 + \mathrm{t}_3} +\mathrm{c}_x.
    \end{aligned}
\end{equation}

{By combing the two equations and eliminating the variable $x_t$, we obtain
\begin{equation}
\begin{aligned}
    \mathrm{u}_s^\mathrm{gt} &= \frac{\mathrm{d}_t^\mathrm{gt}(\mathrm{u}_t - c_x)}{\mathrm{d}_t^\mathrm{gt}+\mathrm{t}_3+\Delta \mathrm{t}_3} + c_x.
\end{aligned}
\end{equation}
%where $\Delta \mathrm{t}_3$ is the motion of the dynamic object. 

\noindent \textbf{Analysis of existing self-supervised depth estimation model.} To minimize the photometric loss, the network ({\ie} DepthNet) will be optimized to encourage the estimated pixel location $\mathrm{u}_s$ to approach the observed pixel location $\mathrm{u}_s^\mathrm{gt}$:
\begin{equation}
\begin{aligned}
   \frac{\mathrm{d}_t(\mathrm{u}_t - c_x)}{\mathrm{d}_t + \mathrm{t}_3} + c_x  &\rightarrow \frac{\mathrm{d}_t^\mathrm{gt}(\mathrm{u}_t - c_x)}{\mathrm{d}_t^\mathrm{gt}+\mathrm{t}_3+\Delta \mathrm{t}_3} + c_x,
\end{aligned}
\end{equation}
This is equivalent to 
\begin{equation}
\begin{aligned}
    \label{eq:simplified_supervision}
    \mathrm{d}_t &\rightarrow \frac{\mathrm{t}_3}{\mathrm{t}_3 + \Delta \mathrm{t}_3}\cdot\mathrm{d}_t^\mathrm{gt}.
\end{aligned}
\end{equation}} 
Here we study four motion patterns of the object and their impacts on DepthNet: static (Figure~\ref{fig:black_hole_theory}: Static scene), movement in the opposite direction of the ego car (Figure~\ref{fig:black_hole_theory}: Dynamic scene (a)), movement in the same direction of the ego car~(Figure~\ref{fig:black_hole_theory}: Dynamic scene (b) and (c)). 

\begin{figure*}
	\begin{center}
	\includegraphics[width=0.92\linewidth]{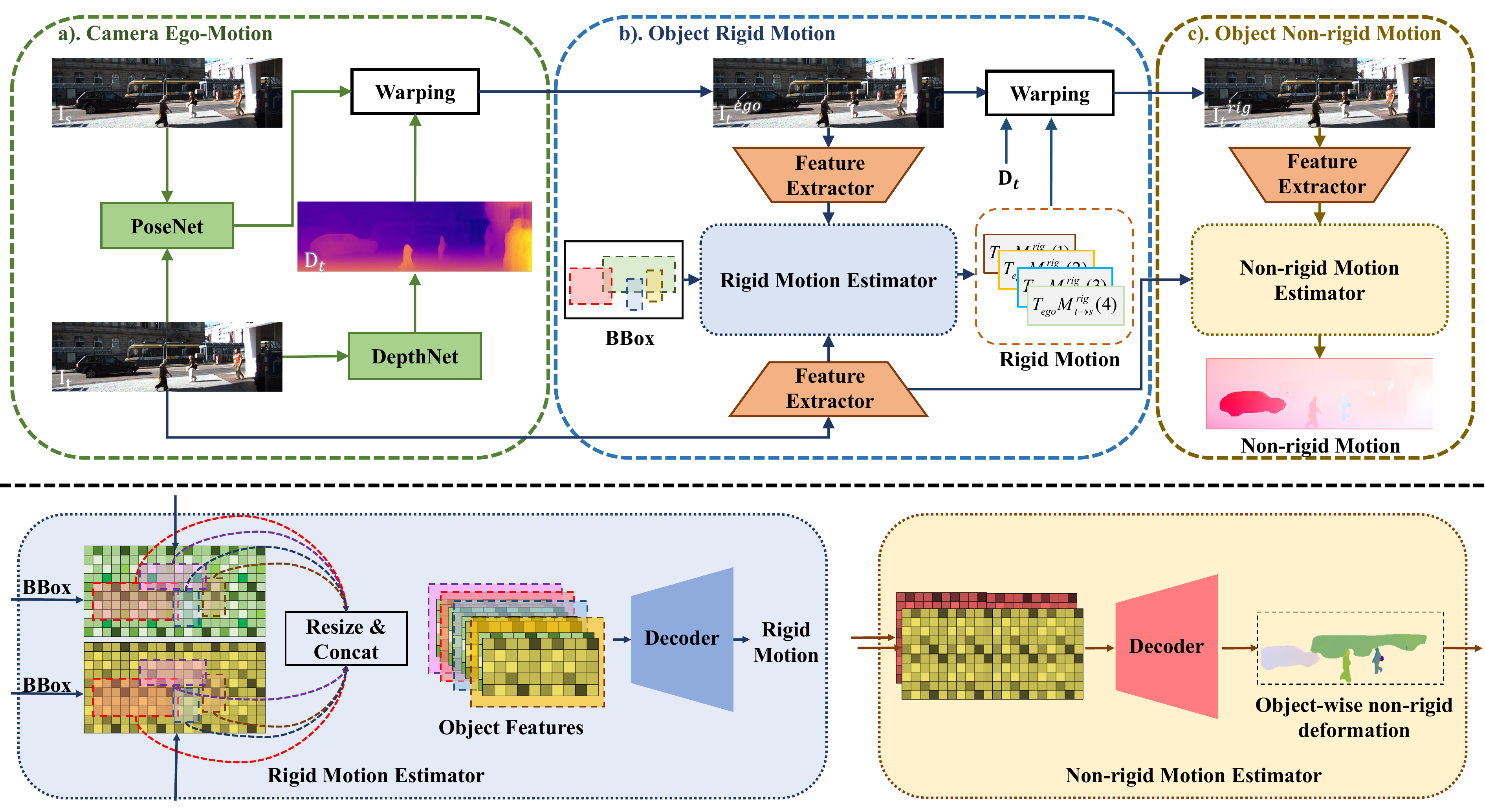}
        \caption{Model overview. Our system requires two consecutive video frames for camera ego-motion prediction (a). The reconstructed image $\mathrm{I}_t^\mathrm{ego}$ and original image $\mathrm{I}_t$ are used as inputs to the dynamic rigid motion module (b) which learns object-wise rigid motion $\mathrm{M}_{t\rightarrow s}^\mathrm{rig}$. Further, the residual non-rigid deformation module (c) exploits $\mathrm{I}_t^\mathrm{rig}$ and $\mathrm{I}_t$ to recover non-rigid deformation $\mathrm{M}_{t\rightarrow s}^\mathrm{def}$. In the rigid motion estimation module, object-wise bounding boxes are obtained by Mask RCNN \cite{he2017mask}. We employ the RoI Align operation~\cite{he2017mask} to generate object-wise features.}
	\label{fig:over_view}
	\end{center}
\end{figure*}

\noindent \textit{Static scene.} 
{This scenario follows the static scene assumption, {\ie} $\Delta \mathrm{t}_3=0$. Hence, according to Equation~\eqref{eq:simplified_supervision},  $\mathrm{d}_t \rightarrow \mathrm{d}_t^{\mathrm{gt}}$~(see Figure~\ref{fig:black_hole_theory}: Static Scene). 
Therefore, minimizing the photometric loss will encourage the model to make correct predictions.}

\noindent \textit{Dynamic scene~(a).}
In this scenario, the car is moving in the opposite direction as the ego car with $\Delta \mathrm{t}_3<0$~(see Figure~\ref{fig:black_hole_theory} (a)). According to Equation~\eqref{eq:simplified_supervision}, we can obtain 
\begin{equation}
\mathrm{d}_t \rightarrow  \frac{\mathrm{t}_3}{\mathrm{t}_3 + \Delta \mathrm{t}_3}\cdot\mathrm{d}_t^\mathrm{gt} < \mathrm{d}_t^\mathrm{gt}
\end{equation}
Thus, minimizing the photometric loss will force DepthNet to predict a smaller depth value in comparison with the ground truth one.

\noindent \textit{Dynamic scene~(b).} 
In this scenario, the car is moving in the same direction as the ego car with movement $\Delta\mathrm{t}_3 \in (0, -\mathrm{t}_3)$~(see Figure~\ref{fig:black_hole_theory} (b)). As in Equation~\eqref{eq:simplified_supervision}, we can conclude that $\mathrm{d}_t$ will monotonically increase as $\Delta\mathrm{t}_3$ increases as  
\begin{equation}
    \begin{aligned}
    \mathrm{d}_t \rightarrow \frac{\mathrm{t}_3}{\mathrm{t}_3 + \Delta \mathrm{t}_3}\cdot\mathrm{d}_t^\mathrm{gt} \in (\mathrm{d}_t^\mathrm{gt}, +\infty).
    \end{aligned}
\end{equation}
If $\Delta\mathrm{t}_3$ approaches $-\mathrm{t}_3$, minimizing the photometric loss will enforce DepthNet to predict an infinite value, {\eg} the black-hole issue.

\noindent \textit{Dynamic scene~(c).}
In this scenario, the object moves in the same direction as the ego car with movement $\Delta\mathrm{t}_3 \in (-\mathrm{t}_3, +\infty)$~(see Figure~\ref{fig:black_hole_theory} (c)). Under this condition, we can obtain that $\mathrm{u}_t < \mathrm{u}_s^\mathrm{gt}$ as $\Delta \mathrm{t}_3 + t_3 >0$. In contrast, according to our previous analysis in ``self-supervised model under static scene assumption'', the estimated source pixel coordinate $\mathrm{u}_s$ will not be smaller than $\mathrm{u}_t$ ($\mathrm{u}_s \geq \mathrm{u}_t$). This means no matter how  DepthNet is optimized, $\mathrm{u}_s$ cannot reach $\mathrm{u}_s^{\mathrm{gt}}$ and  $\mathrm{u}_s\geq \mathrm{u}_t> \mathrm{u}_s^\mathrm{gt}$. Minimizing the photometric loss will encourage DepthNet to produce a large value, approaching infinity ({\ie} the black hole issue), such that $\mathrm{u}_s$ will approach $\mathrm{u}_t$ as Equation~\eqref{eq:limit_u}.

\subsection{Unresolved Research Challenges}\label{sec:unresolved}
The above analysis manifests that without explicitly modeling the dynamics of the moving objects, DepthNet can be optimized in an incorrect direction in dynamic scenes if ${L}_{ph}$ in equation~\eqref{eqn:photo_metric_loss} is adopted during training, leading to erroneous depth predictions for dynamic objects as also verified by our visualization in Figure~\ref{fig:depth_loss_curve}. 

Existing methods attempting to address the dynamic scene issue either estimate 2D residual flow~\cite{yin2018geonet,ranjan2019competitive} or 3D motion~\cite{casser2019struct2depth, cao2019learning, li2020unsupervised, lee2019instance} to alleviate the influence of dynamic objects and, at the same time, be able to estimate motion. However, the fact that motion happens in 3D renders 2D optical flow-based solutions ineffective in reflecting real-world dynamics and makes motion estimation complicated, {\eg} a simple 3D motion of moving forward will lead to complicated patterns in 2D optical flow. The other stream of work, incorporating 3D motion, holds more chances to uncover the real-world dynamic model. However, we find that a dedicated method for estimating 3D motion is still lacking. Existing methods either adopt an object-wise rigid motion model~\cite{lee2019instance,casser2019struct2depth}, which fails in dealing with non-rigid motion or choose to estimate pixel-wise 3D motion~\cite{cao2019learning, li2020unsupervised} which cannot effectively enforce object-level regularization, leading to inconsistent or unreasonable motion predictions of the same object.

In a nutshell, how to model underlying real-world 3D motions and incorporate them into the self-supervised learning pipeline for estimating 3D geometry and motion remains a challenging problem. In this paper, we take one step further and propose a new motion decomposition pipeline to model 3D motion and develop a holistic self-supervised framework for jointly reasoning 3D geometry and motion.

\section{Our Model}\label{sec:model}
In the following, we will elaborate on how we leverage the scene geometry model to design a neural network system for depth and motion estimation and how the system is trained using a self-supervised loss. 
{A given monocular video sequence can be regarded as the projection of the whole 3D scene, including the geometry of static background and the motions of dynamic objects, onto a series of camera image planes determined by a camera pose sequence. We aim to utilize the geometric relationships to recover the underlying 3D scene geometry and motion from monocular videos.}
The overview of the system is shown in Figure~\ref{fig:over_view}. The inputs to the system are two consecutive video frames $\mathrm{I}_t$ and $\mathrm{I}_s$, and our model can simultaneously infer both depth and 3D motions. 
The designs of the depth estimation module and motion estimation module are described in the following sections. 

\subsection{Depth Estimation}
\begin{figure}
    \centering
    \includegraphics[width=1.0\linewidth]{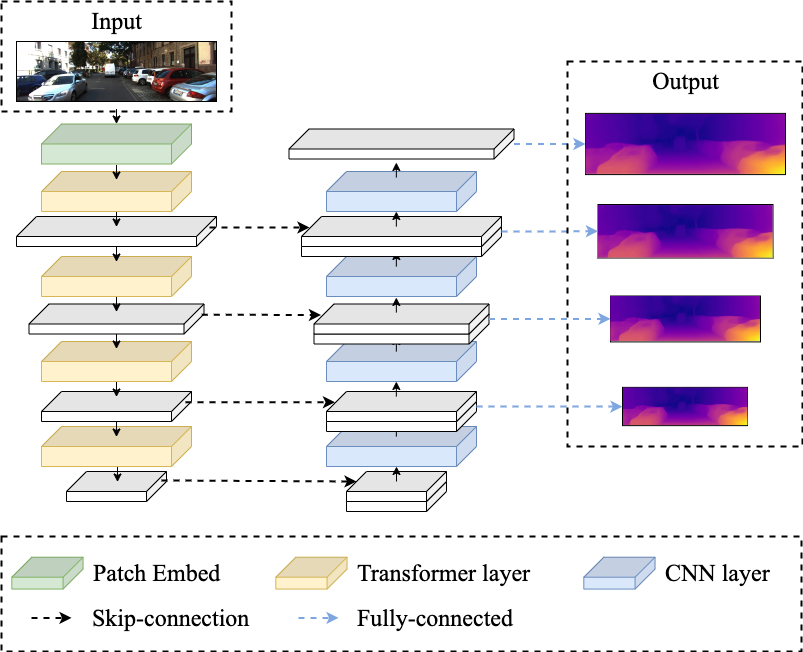}
    % \vspace{-0.6cm}
    \caption{{Network architecture of DepthNet. We propose a hybrid Transformer and CNN model. The encoder consists of four Transformer layers, while the decoder employs four CNN layers. Skip-connections are also introduced for feature fusion in the decoder. DepthNet predicts one depth map given an input frame in a single forward pass. }}
    \label{fig:depthnet}
\end{figure}

{DepthNet is designed to process one frame at a time and produce a depth map $\mathrm{D}_t$ for each input frame $\mathrm{I}_t$.
We build a hybrid Transformer-and-CNN baseline model, as shown in Figure~\ref{fig:depthnet}. 
The Transformer encoder inspired by ~\cite{xie2021segformer} is designed to capture global information, while the CNN decoder is to extract local information. 
Specifically, the input image is split into several patches (a.k.a tokens), and the hierarchical transformer-based encoder architecture is developed to process the input which effectively enlarges the receptive field of the network and allows to capture global geometry clues (see Figure~\ref{fig:depthnet}: Transformer Layer). 
The {pyramid} structure helps the model extract features at different scales, which will be combined with the intermediate features from the decoder. 
Then, the generated feature maps with various scales will be fed to a CNN-based decoder to enhance details by exploiting local information.
We add skip-connection operations to refine features at different scales (see Figure~\ref{fig:depthnet}: CNN Layer).
We also tried to apply the MLP-based design following~\cite{xie2021segformer}, which, however, leads to blurry results. 
In contrast, our hybrid transformer and CNN design can effectively capture global information and refine details for dense predictions.}

\subsection{Motion Estimation} 
\label{sec:motion_estimation}
As shown in Figure~\ref{fig:over_view}, the motion estimation module contains a pose estimation network, PoseNet, to estimate camera ego-motion $\mathrm{T}_{t\rightarrow s}$, and an object motion prediction network, MotionNet (including object rigid motion and object non-rigid motion), to produce object-wise motion map $\mathrm{M}_{t\rightarrow s}$. 

PoseNet (see Figure~\ref{fig:over_view} (a)) is a convolutional neural network with fully connected layers to output the 6-DoF parameters, including Pitch, Roll, Yaw, and translations along three directions. Then, the 6-DoF parameters are converted to a transformation matrix representing camera-ego-motion $\mathrm{T}_{t \rightarrow s}$. 
Given the estimated camera-ego-motion $\mathrm{T}_{t\rightarrow s}$ from PoseNet, we first obtain $\mathrm{I}_t^\mathrm{ego}$ by sampling the source image $\mathrm{I}_s$ according to $\mathrm{p}_{t\rightarrow s}$. The process is detailed in Section~\ref{sec:pre} while $\mathrm{p}_{t\rightarrow s}$ is computed with only $\mathrm{T}_{t\rightarrow s}$ (\emph {i.e.}, object motion $\mathrm{M}_{t\rightarrow s}$ is zero). This process transfers the source image $\mathrm{I}_s$ into the camera coordinate system 
$\mathrm{I}_{t}$ to eliminate the motion caused by camera movements, facilitating the follow-up dynamic object motion estimation process.

To relieve the difficulties of estimating complicated motion patterns for various objects, MotionNet is designed to contain two components: an object-wise rigid motion predictor (see Figure~\ref{fig:over_view} (b)) and {an object-wise non-rigid deformation predictor}~(see {Figure}~\ref{fig:over_view} (c)). 
The object-wise rigid motion predictor takes $\mathrm{I}_t$ and $I_{t}^\mathrm{ego}$ as inputs and then outputs a 6-DoF rigid transformation $\mathrm{M}_{t\rightarrow s}^\mathrm{rig}(i)$ for each object instances $i$ to effectively model rigid motion and produce a reasonably good initialization for modeling non-rigid deformation. For simplicity and consistency of notation, we assign the estimated object-wise rigid motion to the corresponding pixels and thus obtain a rigid motion map denoted as $\mathrm{M}_{t\rightarrow s}^\mathrm{rig}$. Further, {the object-wise non-rigid deformation predictor} is designed to learn a deformation map $\mathrm{M}_{t\rightarrow s}^\mathrm{def}$ by refining the rigid object-wise motion. Thus, the object motion is obtained by combining the rigid motion and non-rigid deformation as 

\begin{equation}
    \mathrm{M}_{t\rightarrow s} = \mathrm{M}_{t\rightarrow s}^\mathrm{rig}(\mathrm{P}_t + \mathrm{M}_{t\rightarrow s}^\mathrm{def}) - \mathrm{P}_t.
    \label{eq:obj_motion}
\end{equation}

\begin{figure}[t]
	\begin{center}
		\includegraphics[width=\linewidth]{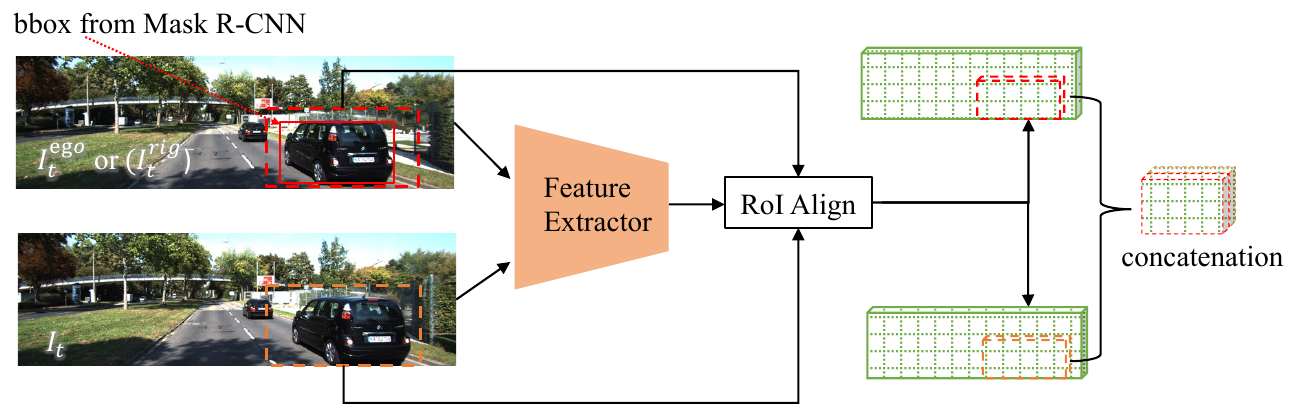}
	\end{center}
	\caption{{We utilize a feature extractor~(\ie ResNet18) to extract feature maps for two given input frames. Bounding boxes obtained from a pre-trained Mask R-CNN are applied to crop the features of the RoI area using the RoI Align operation. These features are then concatenated and finally sent to downstream tasks~(\ie object motion estimation).}}
	\label{fig:roi_align}
\end{figure}

\noindent\textbf{Dynamic rigid motion estimation.}
The Rigid Motion Estimator takes as inputs the image features within the object area of both the reconstructed image $\mathrm{I}_t^\mathrm{ego}$ and the original image $\mathrm{I}_t$, as shown in Figure \ref{fig:over_view}. Its objective is to estimate a 6-DoF rigid transformation matrix $\mathrm{M}_{t\rightarrow s}^\mathrm{rig} (i)$ for each moving object $i$.
It is well-noted that we do not employ the ground-truth semantic mask on the KITTI dataset or the Cityscapes dataset to separate objects from the background. Instead, similar to prior studies \cite{lee2021learning,casser2019struct2depth,guizilini2020semantically}, we generate a semantic mask using a pre-trained instance segmentation framework Mask R-CNN~\cite{he2017mask}. The Mask R-CNN is pre-trained on the COCO dataset, which is available on their official  \href{https://github.com/facebookresearch/detectron2/}{GitHub website}. This further enhances the generalizability of our approach. We utilize all the instance masks output by the pre-trained instance segmentation network, but upon visualizing the generated mask, we notice that the majority represents vehicles and pedestrians.
The dynamic rigid motion network employs several convolutional layers to extract feature representations separately given $\mathrm{I}_{t}^\mathrm{ego}$ and $\mathrm{I}_t$~(see Figure~\ref{fig:over_view}: Rigid Motion Estimator). We use RoI Align~\cite{he2017mask} to extract object-wise features from each encoded feature map. The original object bounding box prediction is enlarged by 20 pixels to incorporate context information~(see Figure~\ref{fig:roi_align} for more details). The extracted RoI features from $\mathrm{I}_t$ and $\mathrm{I}_t^\mathrm{ego}$ are concatenated and then fed into a network with several convolutional layers followed by a fully connected layer to output the 6-DoF rigid motion $\mathrm{M}_{t\rightarrow s}^\mathrm{rig} (i)$ for each object instance $i$. In this stage, we remove objects that have already been well reconstructed in $\mathrm{I}_t^\mathrm{rig}$ with camera ego-motion, {\ie} static objects. Specifically, we compare the photometric loss in object areas before/after applying our object rigid motion estimation module. If the loss increases, there is a high chance that the object is static and is already well reconstructed in the previous stage. 

\noindent\textbf{Non-rigid deformation.}
Given the estimated object-wise 6-DoF rigid transformation $\mathrm{M}_{t\rightarrow s}^\mathrm{rig}(i)$, $\mathrm{I}_t^\mathrm{rig}$ is obtained by sampling from $\mathrm{I}_s$ through transforming each instance separately. 
This is achieved by computing $\mathrm{p}_{t\rightarrow s}$ considering $\mathrm{D}_t$, $\mathrm{T}_{t\rightarrow s}$, and $\mathrm{M}_{t\rightarrow s}$ with $\mathrm{M}_{t\rightarrow s}^\mathrm{def}$ being zero as Equation~\eqref{eq:inverse_warpping_rigid}. Then, $\mathrm{I}_t^\mathrm{rig}$ is obtained using inverse warping following~\cite{garg2016unsupervised, zhou2017unsupervised, godard2019digging}.
\begin{equation}
    \mathrm{d}_s\mathrm{p}_s = \mathrm{KT}_{t\rightarrow s}(\mathrm{d}_t \mathrm{K}^{-1} \mathrm{p}_t + \mathrm{M}_{t\rightarrow s}(\mathrm{p}_t)).
	\label{eq:inverse_warpping_rigid} 
\end{equation}
The formation of $\mathrm{I}_t^\mathrm{rig}$ considers camera-ego-motion and object-wise rigid motion and eliminates their effects, which is further used as an input for a pixel-wise non-rigid deformation module. As shown in Figure~\ref{fig:over_view}, the feature extractor of the previous stage is reused to encode inputs $\mathrm{I}_t$ and $\mathrm{I}_t^\mathrm{rig}$. The output features are concatenated together and fed into a pixel-wise motion deformation estimator. The estimator contains several convolution and up-sample layers and outputs a pixel-wise deformation map with three channels representing motion in the $x, y,$ and $z$ directions.
Note that we also remove objects that have already been well reconstructed in $\mathrm{I}_t^\mathrm{rig}$ with camera ego-motion and object-wise rigid motion, which have a high potential of being static objects or objects with rigid motion.

\begin{figure*}
    \begin{center}
        \includegraphics[width=0.95\linewidth]{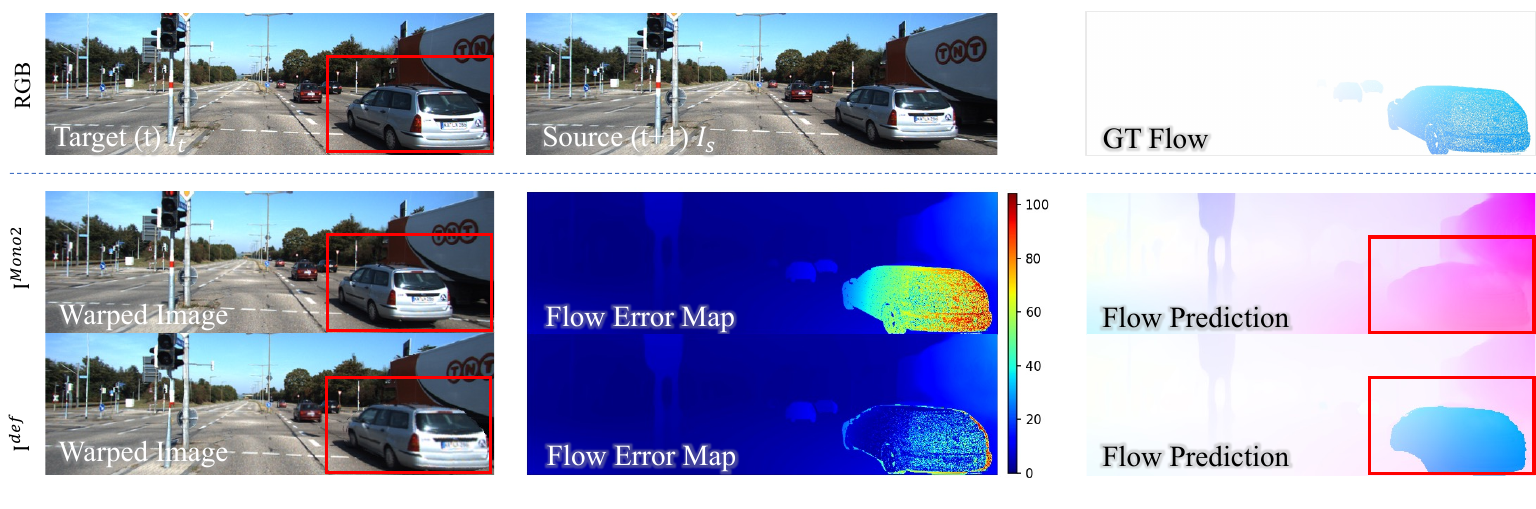}
    \end{center}
    
    \caption{Qualitative comparisons with MonoDepth2~\cite{godard2019digging}. $\mathrm{I}^{Mono2}$ stands for the warped image generated by MonoDepth2~\cite{godard2019digging}. Our method revises dynamic foreground optical flow estimation through motion decomposition modules. Red boxes illustrate the major differences.
    The ground truth flow provided is shown on the upper right side.}
    \label{fig:kitti_flow_rigid_obj_gt_epe}
\end{figure*}

\subsection{Self-supervised Learning} 
In the following, we show how we can train the network in a self-supervised manner. Based on the estimated depth map $\mathrm{D}_t$, camera ego-motion $\mathrm{T}_{t\rightarrow s}$, and the object motion $\mathrm{M}_{t\rightarrow s}$, we can calculate the pixel correspondence $p_{t\rightarrow s}$ between target frame and source frame using {Equation~\eqref{eq:inverse_warpping_rigid}}. 
Based on this correspondence, the target frame is reconstructed by sampling the corresponding pixels from the source frame $\mathrm{I}_s$. As before, the reconstructed target frame $\hat{\mathrm{I}}_t$  can be obtained by utilizing bi-linear interpolation following~\cite{zhou2017unsupervised, jaderberg2015spatial}.  
$\hat{\mathrm{I}}_t$ can be $\mathrm{I}_t^\mathrm{ego}$, $\mathrm{I}_t^\mathrm{rig}$ and $\mathrm{I}_t^\mathrm{def}$ which are reconstructed frames. Our self-supervised objective is built by comparing the reconstructed frame $\hat{\mathrm{I}}_t$ with the observed frame $\mathrm{I}_t$. 

We use the photometric loss as  Equation~\eqref{eqn:photo_metric_loss} to measure the discrepancy between $\hat{\mathrm{I}}_t$ and $\mathrm{I}_t$. It encourages the network to learn to estimate $\mathrm{D}_t$, $\mathrm{T}_{t\rightarrow s}$ and $\mathrm{M}_{t\rightarrow s}$ ($\mathrm{M}_{t\rightarrow s}^{\mathrm{rig}} (i)$ for all instances $i$ and $\mathrm{M}_{t\rightarrow s}^{\mathrm{def}}$) that can produce $\mathrm{I}_t$ given $\mathrm{I}_s$. 
Furthermore, we also incorporate a smoothness loss $\mathcal{L}_\mathrm{ds}$ on the produced depth map to encourage local smoothness: 
\begin{equation}
	\mathcal{L}_\mathrm{ds} = |\mathrm{\partial}_x \mathrm{D}^*_{t}|e^{-|\mathrm{\partial}_x \mathrm{I}_{t}|} + |\mathrm{\partial}_y \mathrm{D}^*_{t}|e^{-|\mathrm{\partial}_y \mathrm{I}_{t}|},
	\label{eqn:ds_loss}
\end{equation}

\noindent where $\mathrm{D}_t^*=\mathrm{D}_t/\overline{\mathrm{D}_t}$ is the mean-normalized inverse depth map following~\cite{godard2019digging}. We also introduce a mask loss $\mathcal{L}_m$ to regularize the reconstruction in the semantic space, which is defined as:

\begin{equation}
	\mathcal L_{m} = 1 - \mathrm{IoU}(\mathrm{\hat{m}_t}, \mathrm{m}_{t}),
\end{equation}

\noindent where $\hat{m}_t$ represents the reconstructed semantic mask according to the estimated geometric model, and $\mathrm{IoU}(\cdot, \cdot)$ computes the Intersection over Union~(IoU) between $\hat{m}_t$ and the mask $m_t$ of frame $\mathrm{t}$. 
Finally, $\mathcal{L}_\mathrm{ph}$, $\mathcal{L}_\mathrm{ds}$ and $\mathcal{L}_m$ are combined to be the overall loss:
\begin{equation}
	\mathcal{L} = \mathrm{\omega}_\mathrm{ph} \cdot \mathcal{L}_\mathrm{ph} + \mathrm{\omega}_\text{ds} \cdot \mathcal{L}_\mathrm{ds} + \mathrm{\omega}_\mathrm m \cdot \mathcal{L}_\mathrm{m},
	\label{eqn:total_loss} 
\end{equation}

\noindent where hyper-parameters $\omega_\mathrm{ph},~\omega_\mathrm{ds}$, and $\omega_m$ stand for the loss weight of the photometric loss, smoothness loss, and mask loss, respectively. $\omega_\mathrm{ph}$,  $\omega_\mathrm{ds}$, and $\omega_{m}$ are set to $1.0$, $0.001$, and $1.0$, respectively.

The whole framework is trained in a piece-wise manner: 
1) Train DepthNet and PoseNet using $\mathrm{I}_s$ and $\mathrm{I}_{t}$;
2) Fix DepthNet and PoseNet from stage 1, and train object-wise rigid motion estimator together with feature extractor using $\mathrm{I}_t^\mathrm{ego}$ and $\mathrm{I}_t$;
3) Fix DepthNet, PoseNet, feature extractor, and object-wise rigid motion estimator, and train pixel-wise motion deformation estimator using $\mathrm{I}_t^\mathrm{rig}$ and $\mathrm{I}_t$. As all the components are differentiable, the network can be optimized by minimizing the self-supervised loss;
4) Fix feature extractor, object-wise rigid motion estimator, and pixel-wise motion deformation estimator, and finetune DepthNet and PoseNet using $\mathrm{I}_t^\mathrm{def}$ and $\mathrm{I}_t$. 

\section{Experiments}\label{sec:experiments}
In this section, we will first describe the implementation details of our experiments in Section~\ref{sec:implementation} and then introduce the datasets and evaluation metrics in Section~\ref{sec:datasets} and Section~\ref{sec:metrics}. Quantitative and qualitative results are presented in Section~\ref{sec:main_results}.

\subsection{Implementation Details}\label{sec:implementation}
The network undergoes training on a single NVIDIA 3090 GPU, utilizing the Adam optimizer \cite{kingma2014adam} with hyperparameters $\beta_1=0.9$ and $\beta_2=0.999$. By default, all RGB inputs are resized to 640 $\times$ 192 unless otherwise specified. DepthNet and PoseNet are both trained using a learning rate of $5 \times 10^{-5}$, while MotionNet's training employs a learning rate of $10^{-5}$.
% As described in paragraph \textbf{Dynamic rigid motion estimation} and \textbf{Non-rigid deformation} in Section \ref{sec:motion_estimation}, before training, we employ a reconstruction loss to selectively identify dynamic scenes, enhancing training efficiency. This design only results in a small increase in training time. For instance, when training on the Cityscapes dataset, our recently introduced motion decomposition module only takes approximately 4 hours, which represents a relatively minor increase in training costs when compared to the original training requirements for DepthNet and PoseNet, which takes 20-30 hours.

\subsection{Datasets}\label{sec:datasets}
We conduct experiments on the KITTI 2015, Cityscapes, and VKITTI2 datasets. The KITTI 2015 dataset consists of image sequences for $200$ driving scenes at $10$ frames per second. The image resolution is around $375 \times 1242$. Nevertheless, motion patterns and dynamic objects are still limited in the KITTI dataset. Thus, we also evaluate our model on the Cityscapes and VKITTI2 datasets that cover dense and dynamic scenarios. The original frame resolutions in the Cityscapes and VKITTI2 datasets are around $1024 \times 2048$ and $375 \times 1242$, respectively.

\noindent\textbf{KITTI Eigen depth split.}
We use the Eigen depth split~\cite{eigen2015predicting} in the depth estimation task.
Following~\cite{godard2019digging}, the invalid static frames are removed in the pre-processing step. 
Finally, the train, validation, and test set contain $39810$, $4424$, and $673$ images with resolutions around $375 \times 1242$, respectively.

\noindent\textbf{Cityscapes split.}
We split the Cityscapes dataset into train split and test split following Manydepth~\cite{watson2021temporal}, which consists of $69731$ and $1525$ images. 
{Training images are all collected from the monocular sequences so that we can use the neighboring frames for training.} 
Cityscapes split is employed for depth estimation, and we use the provided disparity maps from SGM~\cite{hirschmuller2007stereo} for evaluation.

\noindent\textbf{VKITTI2 split.}
As existing monocular methods are not evaluated on the VKITTI2 dataset~\cite{cabon2020vkitti2, gaidon2016virtual}, we randomly split the dataset into two subsets~(train split and test split), which consist of $1904$ and $212$ images, respectively. 
It is notable that VKITTI2 provides ground truth depth. Thus, we can conduct depth estimation evaluation on VKITTI2.

\noindent\textbf{KITTI 2015 optical flow train split.}
To better evaluate the predicted motion, we also adopt the KITTI 2015 optical flow split and employ its training split~(200 images) to validate our model in the motion estimation task.

\subsection{Evaluation metrics}\label{sec:metrics}
\noindent\textbf{Depth estimation.}
We follow the depth evaluation metrics in~\cite{godard2017unsupervised}: absolute relative error~(Abs Rel), square relative error~(Sq Rel), root mean square error~(RMSE), and root mean square logarithmic error~(RMSE log). 
{``$\delta < \sigma$'' represents the accuracy of depth predictions, \ie the percentage of $\mathrm{d}_i$ satisfying $\mathrm{max}(\frac{d_i}{d_i^\mathrm{gt}},\frac{d_i^\mathrm{gt}}{d_i})<\sigma$.}
To eliminate the influence of scale which is missing in the monocular scenario, Godard \etal~\cite{godard2017unsupervised} scale the predicted depth frame by the median of the ground truth in evaluation. 

In addition, the predicted monocular depth is capped at 80m during evaluation following~\cite{godard2019digging}. We evaluate our model in foreground and background regions to assess whether our model can benefit the depth estimation of dynamic objects. The foreground region includes all the dynamic objects but excludes the background region. 
Hence, the performance in the foreground region can better reflect the performance of dynamic objects in comparison with the overall evaluation metric. The foreground and background are split by the ground truth instance masks. 
If the ground truth instance masks are not provided, we use Mask R-CNN~\cite{he2017mask} to generate them.

\noindent\textbf{Optical flow/Scene flow estimation.}
For the optical flow estimation task, we use the average endpoint error~(EPE) metric~\cite{baker2011database} in both ``Noc'' and ``Occ'' regions. ``Noc'' stands for non-occluded regions, and ``Occ'' indicates all the regions, including both occluded and non-occluded regions. To better assess the ability for motion estimation of dynamic objects, we divide an image into foreground and background regions and evaluate our model in these two regions separately. To test the performance of scene flow estimation, we also apply the metrics adopted in the KITTI 2015 Scene Flow benchmark~\cite{menze2015joint, menze2015object}. The metrics report the outlier percentage of the estimated disparity for the target frame~(D0) and for the source frame mapped into the target frame~(D1). In addition, F1 computes the proportion of outliers in the optical flow estimation task. Specifically, outliers are defined as pixels whose estimation errors either exceed 3 pixels or 5\% w.r.t. the ground truth disparity or optical flow. Moreover, SF is the metric for scene flow estimation, which shows the outlier percentage of the pixels that are considered outliers in any of these three metrics(\ie D0, D1, and F1).

\subsection{Main results}~\label{sec:main_results}
% We conduct a comprehensive evaluation of our approach by comparing it with state-of-the-art methods in monocular depth estimation, optical flow estimation, and scene flow estimation. It is worth noting that, to the best of our knowledge, there are no existing works that have been evaluated on the VKITTI2 dataset and DAVIS dataset. Therefore, we present the quantitative results obtained on the VKITTI2 dataset and DAVIS dataset and perform a detailed analysis focusing on background and foreground areas in Section \ref{sec:ablation}.
We conduct a comprehensive evaluation of our approach by comparing it with state-of-the-art methods in monocular depth estimation, optical flow estimation, and scene flow estimation. It is worth noting that, to the best of our knowledge, there are no existing works that have been evaluated on the VKITTI2 dataset. Therefore, we present the quantitative results obtained on the VKITTI2 dataset and perform a detailed analysis focusing on background and foreground areas in Section \ref{sec:ablation}.

\subsubsection{Monocular Depth Evaluation}\label{sec:depth_evaluation}
\noindent\textbf{Results on KITTI eigen depth split.}
The depth estimation results, along with comparisons to state-of-the-art (SOTA) methods, are presented in Table~\ref{tab:kitti_depth}. To ensure a fair comparison, given that existing works train on images of varying sizes, we evaluate our algorithms across different resolution settings. Our model demonstrates superior performance in all tested configurations. Specifically, it establishes new SOTA benchmarks in the majority of resolutions (416$\times$128, 1024$\times$320, and 1280$\times$384). Notably, while recent studies, such as the one by Guizilini \etal \cite{guizilini2020semantically}, leverage additional training data to enhance results, our model achieves better overall performance using only the KITTI dataset.

\begin{table*}[!htbp]
\centering
\caption{Quantitative results of monocular depth estimation on the KITTI Eigen Depth Split~\cite{geiger2012we}. K, CS, and VK stand for the KITTI, Cityscapes, and VKITTI datasets, respectively. Bold font indicates the best result.}
\scalebox{0.56}{
    % \begin{tabular}{l|c|c|c|c|c|c|c|c|c}
    \begin{tabular}{lccccccccc}
    \toprule  
    Methods & Datasets & Size & Abs Rel $\downarrow$ & Sq Rel $\downarrow$ & RMSE $\downarrow$ & RMSE log $\downarrow$ & $\delta < 1.25 \uparrow$ & $\delta < 1.25^2 \uparrow$ & $\delta < 1.25^3 \uparrow$ \\ \hline
    % \hline
    Pilzer \etal~\cite{pilzer2018unsupervised} & K &  $512 \times 256$ & 0.152 & 1.388 & 6.016 & 0.247 & 0.789 & 0.918 & 0.965 \\
    GeoNet~\cite{yin2018geonet} & CS+K  &  $416\times128$ & 0.149 & 1.060 & 5.567 & 0.226 & 0.796 & 0.935 & 0.975 \\
    Struct2Depth~\cite{casser2019struct2depth} & K & $416 \times 128$ & 0.141 & 1.026 & 5.291 & 0.215 & 0.816 & 0.945 & 0.979 \\
    Li {\etal}~\cite{li2020unsupervised} & K & $416 \times 128$ & 0.130 & 0.950 & 5.138 & 0.209 & 0.843 & 0.948 & 0.978 \\
    Gordon\etal~\cite{gordon2019depth} & K & $416 \times 128$ & 0.128 & 0.959 & 5.230 & 0.212 & 0.845 & 0.947 & 0.976 \\
    \textbf{DO3D~(ours)} & K & $416 \times 128$ & \textbf{0.116} & \textbf{0.830} & \textbf{4.964} & \textbf{0.196} & \textbf{0.867} & \textbf{0.957} & \textbf{0.981} \\
    \hline
    SfMLearner~\cite{zhou2017unsupervised} & CS+K  &  $640 \times 192$ & 0.183 & 1.595 & 6.709 & 0.270 & 0.734 & 0.902 & 0.959 \\
    SGDepth~\cite{klingner2020self} & K      &  $640 \times 192$ & {0.113} & {0.835} & {4.693} & {0.191} & {0.879} & {0.961} & 0.981 \\
    Monodepth2~\cite{godard2019digging} & K & $640 \times 192$ & 0.115 & 0.903 & 4.863 & 0.193 & 0.877  & 0.959  & 0.981 \\ 
    PackNet-SFM~\cite{guizilini2019packnet} & K  &  $640 \times 192$ & {0.111} & {0.785} & {4.601} & {0.189} & {0.878} & {0.960} & {0.982} \\
    HR-Depth~\cite{lyu2020hr} & K  &  $640 \times 192$ & 0.109 & 0.792 & 4.632 & 0.185 & 0.884 & 0.962 & 0.983 \\
    Johnston {\etal}~\cite{johnston2020self} & K  &  $640 \times 192$ & {0.106} & {0.861} & {4.699} & {0.185} & {0.889} & {0.962} & {0.982} \\
    MovingSLAM~\cite{xu2021moving} & K   &  $640 \times 192$ & {0.105} & 0.889 & 4.780 & {0.182} & 0.884 & 0.961 & 0.982 \\
    Jung {\etal}~\cite{jung2021fine} & K  &  $640 \times 192$ & {0.105} & {0.722} & {4.547} & {0.182} & {0.886} & \textbf{0.964} & \textbf{0.984} \\
    Guizilini {\etal}~\cite{guizilini2020semantically} & CS+K  &  $640 \times 192$ & \textbf{0.102} & \textbf{0.698} & \textbf{4.381} & \textbf{0.178} & \textbf{0.896} & \textbf{0.964} & \textbf{0.984} \\
    \textbf{DO3D~(ours)} & K & $640 \times 192$ & 0.106 & 0.797 & 4.582 & 0.181 & 0.890 & \textbf{0.964} & 0.983 \\ % stage3d_segformer2b5_v6_pre_lr5_epoch40_new, weights_1
    % TODO: 0.104, node61
    \hline
    EPC++~\cite{luo2019every} & K & $832 \times 256$ & 0.141 & 1.029 & 5.350 & 0.216 & 0.816 & 0.941 & 0.976 \\
    CC~\cite{ranjan2019competitive} & CS+K & $832 \times 256$ & 0.139 & 1.032 & 5.199 & 0.213 & 0.827 & 0.943 & 0.977 \\
    Lee {\etal}~\cite{lee2019instance} & K &  $832 \times 256$ & 0.124 & 1.009 & 5.176  & 0.208 & 0.839  & 0.942 & 0.980 \\ 
    DAM+CSAC~\cite{lee2021attentive} & K     &  $832 \times 256$ & 0.114 & 0.876 & 4.715 & 0.191 & 0.872 & 0.955 & 0.981 \\
    Monodepth2~\cite{godard2019digging} & K & $1024 \times 320$ & 0.115 & 0.882 & 4.701 & 0.190 & 0.879 & 0.961 & 0.982 \\
    HR-Depth~\cite{lyu2020hr} & K & $1024 \times 320$ & 0.106 & 0.755 & 4.472 & 0.181 & 0.892 & 0.966 & \textbf{0.984} \\
    FeatDepth~\cite{shu2020feature}& K & $1024 \times 320$ & 0.104 & 0.729 & 4.481 & 0.179 & 0.893 & 0.965 & \textbf{0.984} \\
    \textbf{DO3D~(ours)} & K & $1024 \times 320$ & \textbf{0.099} & {\textbf{0.697}} & \textbf{4.397} & {\textbf{0.178}} & \textbf{0.901} & {\textbf{0.967}} & 0.983 \\
    \hline
    SGDepth~\cite{klingner2020self} & K & $1280 \times 384$ & 0.113 & 0.880 & 4.695 & 0.192 & 0.884 & 0.961 & 0.981 \\
    Zhou \etal~\cite{zhou2019unsupervised} & K & $1280 \times 384$ & 0.121 & 0.837 & 4.945 & 0.197 & 0.853 & 0.955 & 0.982 \\
    HR-Depth~\cite{lyu2020hr} & K & $1280 \times 384$ & 0.104 & 0.727 & 4.410 & 0.179 & 0.894 & 0.966 & \textbf{0.984} \\
    PackNet-SFM~\cite{guizilini2019packnet} & K & $1280 \times 384$ & 0.107 & 0.802 & 4.538 & 0.186 & 0.889 & 0.962 & 0.981 \\
    Guizilini \etal~\cite{guizilini2020semantically} & CS+K & $1280 \times 384$ & 0.100 & 0.761 & \textbf{4.270} & 0.175 & \textbf{0.902} & 0.965 & 0.982 \\
    % \textbf{DO3D~(ours)} & K & $1280 \times 384$ & \textbf{0.098} & \textbf{0.724} & 4.316 & \textbf{0.174} & \textbf{0.905} & \textbf{0.968} & \textbf{0.984} \\
    % TODO: LOST
    \textbf{DO3D~(ours)} & K & $1280 \times 384$ & \textbf{0.100} & \textbf{0.651} & 4.280 & \textbf{0.173} & 0.897 & \textbf{0.968} & \textbf{0.985} \\
    \bottomrule
    \end{tabular}
}
\label{tab:kitti_depth}
\end{table*}

\begin{figure}
    \begin{center}
        \includegraphics[width=1\linewidth]{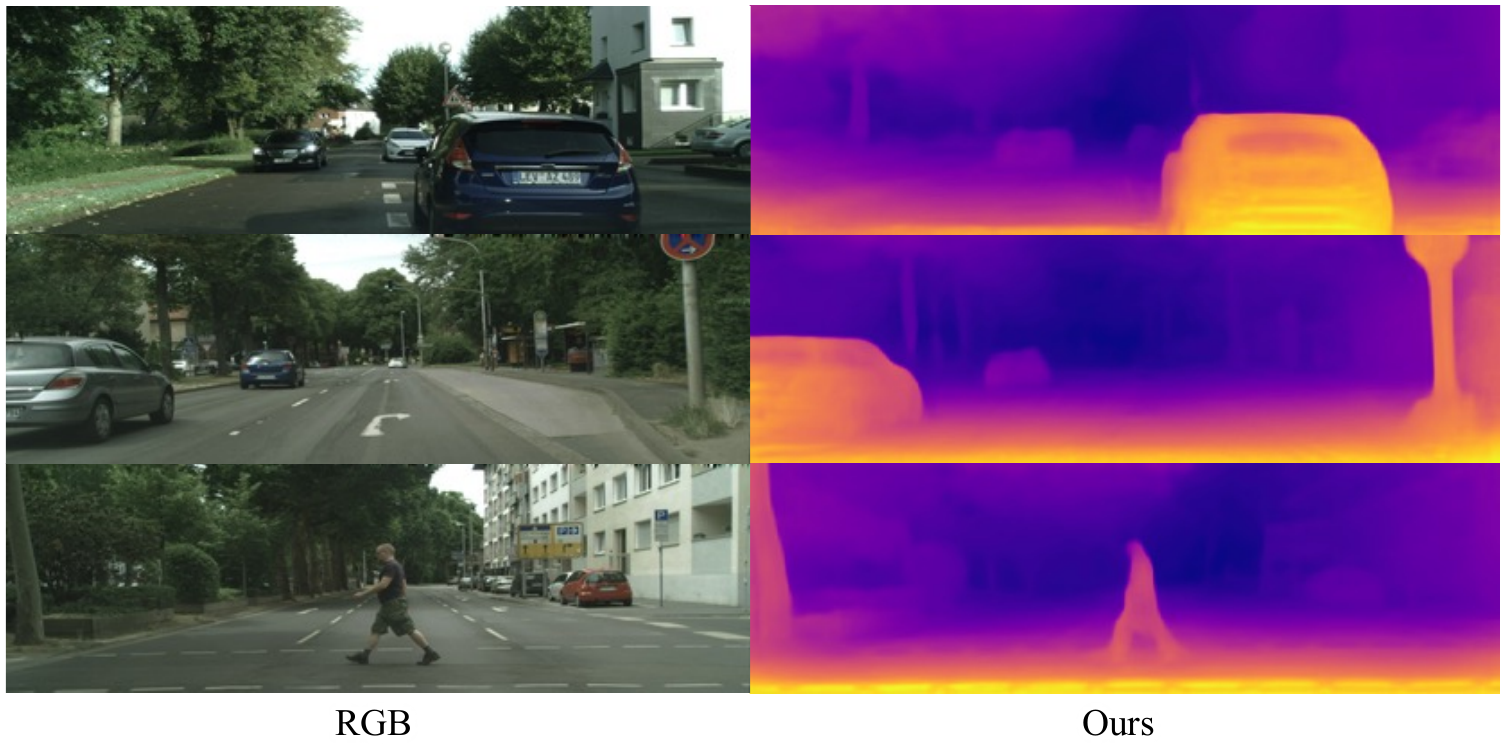}
    \end{center}
    \caption{{Visualization of depth estimation results on the Cityscapes dataset. Three representative scenes are selected: thin things, dynamic objects with rigid motions, and dynamic objects with non-rigid motions.}}
    \label{fig:depth_visualization_cs}
\end{figure}

\noindent\textbf{Results on Cityscapes split.} 
Note that there are no standard training and evaluation settings in the Cityscapes dataset. Previous works either do not provide details about their training sets~\cite{casser2019struct2depth,gordon2019depth} or use $22,973$ image pairs~\cite{pilzer2018unsupervised,li2020unsupervised}, which are widely used in segmentation tasks, but not all of them provide neighboring frames, making them not suitable for our setting. Therefore, we follow the setting in ManyDepth~\cite{watson2021temporal} which collects training data from image sequences. We do not compare with ManyDepth~\cite{watson2021temporal} since they need to employ multiple frames at test time, while our model focuses on monocular depth estimation. Since dynamic scenes are more challenging for self-supervised depth estimation methods, few existing works report their results on this dataset. To make the model converge better, DepthNet and PoseNet are initialized using weights pre-trained on the KITTI dataset. Quantitative results on the Cityscape dataset are shown in Table~\ref{tab:cs_depth}. Our model gains competitive performance on Cityscapes. Besides, our model attains a more significant performance improvement in the foreground areas as in Table~\ref{tab:kitti_depth_fgbg}. This further verifies the effectiveness of our formulation in modeling dynamic scenes. In terms of qualitative results, our model predicts more accurate 3D motion~(see Figure \ref{fig:3d_motion}) and significantly outperforms the other compared works~\cite{casser2019struct2depth, pilzer2018unsupervised, gordon2019depth, li2020unsupervised} on the KITTI dataset. As shown in Figure~\ref{fig:depth_visualization_cs},  our model can well preserve details, which can be seen in the edge areas of thin structures~({\eg} signs and trees). The black-hole issues are alleviated, and our model can predict clear and accurate depth maps for dynamic objects with both rigid and non-rigid motions~(\eg cars and pedestrians).

\begin{table*}[!htbp]
\centering
\caption{Quantitative results of monocular depth estimation on the Cityscapes dataset. Bold font indicates the best result. }
\scalebox{0.69}{
    % \begin{tabular}{l|c|c|c|c|c|c|c|c}
    \begin{tabular}{lcccccccc}
    \toprule  
    Methods & Size & Abs Rel $\downarrow$ & Sq Rel $\downarrow$ & RMSE $\downarrow$ & RMSE log $\downarrow$ & $\delta < 1.25 \uparrow$ & $\delta < 1.25^2 \uparrow$ & $\delta < 1.25^3 \uparrow$ \\ 
    \hline
    % \hline
    Struct2Depth~\cite{casser2019struct2depth} & $416 \times 128$ & 0.145 & 1.737 & 7.280  & 0.205 & 0.813  & 0.942 & 0.976   \\
    Pilzer \etal~\cite{pilzer2018unsupervised} &  $512 \times 256$ & 0.240 & 4.264 & 8.049  & 0.334 & 0.710  & 0.871 & 0.937   \\
    Gordon \etal~\cite{gordon2019depth} &  $416 \times 128$ & {0.127} & 1.330 & {6.960}  & {0.195} & {0.830}  & 0.947 & {0.981}   \\
    Li {\etal}~\cite{li2020unsupervised} & $416 \times 128$ & 0.119 & {1.290} & {6.980}  & {0.190} & 0.846  & 0.952 & 0.982   \\
    % \hline
    % \textbf{DO3D~(ours)} & $416 \times 128$ & 0.121 & \textbf{1.188} & \textbf{6.764} & \textbf{0.181} & \textbf{0.849} & \textbf{0.965} & \textbf{0.990} \\
    \textbf{DO3D~(ours)} & $416 \times 128$ & \textbf{0.100} & \textbf{0.977} & \textbf{6.018} & \textbf{0.156} & \textbf{0.894} & \textbf{0.976} & \textbf{0.992} \\
    \bottomrule
    \end{tabular}
}
\label{tab:cs_depth}
\end{table*}

\begin{figure}
    \begin{center}
        \includegraphics[width=1\linewidth]{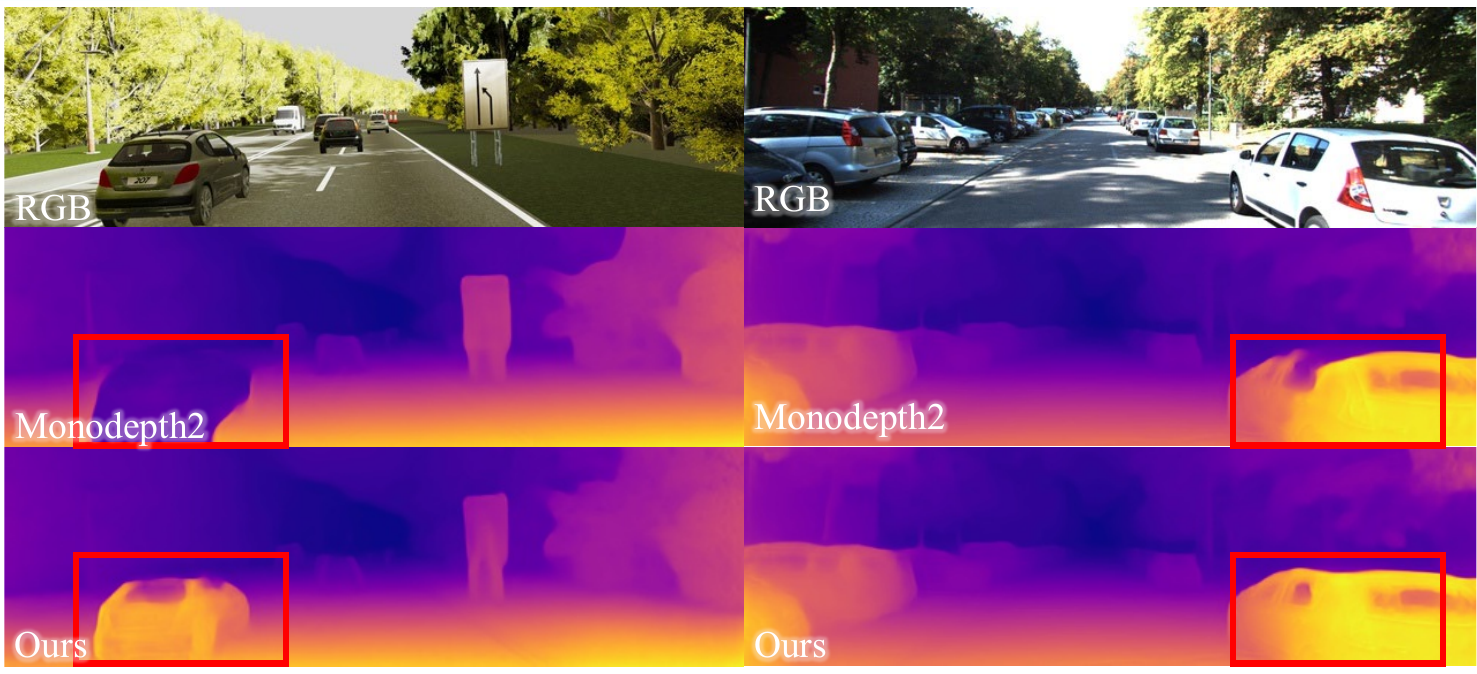}
    \end{center}
    \caption{Depth estimation visualization results on the VKITTI2 dataset~\cite{cabon2020vkitti2}~(left) and KITTI dataset~(right). The middle row shows the depth predicted by Monodepth2~\cite{godard2019digging} and the last row shows the depth predicted by our model, respectively.}
    \label{fig:blackhole}
\end{figure}

\noindent\textbf{More qualitative results.}
In addition to the Cityscapes dataset, qualitative results {on VKITTI2 and KITTI datasets} shown in {Figure}~\ref{fig:blackhole} also demonstrate the superiority of our proposed method, especially at object boundaries. With our motion module to faithfully model dynamic scenes, the depth estimation quality of dynamic objects is improved, and the infinite depth phenomenon in dynamic object areas is alleviated; see Figure~\ref{fig:blackhole} red box. 

\subsubsection{Optical Flow Evaluation}\label{sec:optical_flow_evaluation}
We perform quantitative and qualitative analyses on the KITTI 2015 Optical Flow Train Split to evaluate the effectiveness of our 2D motion estimation, also referred to as optical flow estimation. Our comparisons include not just models that are representative of geometry-based optical flow estimation approaches, such as GeoNet~\cite{yin2018geonet} and Monodepth2~\cite{godard2019digging}, but also methods that focus on pixel matching~\cite{ren2017unsupervised, ranjan2019competitive, hur2020self, zou2018df}. 
Note that geometry-based optical flow estimation, which estimates flow based on depth, camera ego-motion, and object motion, is more challenging than the pixel-matching-based approach. This is because the latter can explicitly leverage texture and context similarities of nearby frames to obtain optical flow while the former has to estimate optical flow in an implicit manner.  Also, pure matching-based methods cannot generate 3D scene flow for comparison. Therefore, in the comparison, we only include matching-based methods~\cite{yin2018geonet,zou2018df,ranjan2019competitive,godard2019digging} that are developed for self-supervised motion and depth estimation.
The results are shown in Table~\ref{tab:kitti_sceneflow_opticalflow}. 
We can conclude that our model achieves the best overall performance in both occluded and non-occluded regions, demonstrating the effectiveness of our approach.
Qualitative results are shown in Figure~\ref{fig:kitti_flow_rigid_obj_gt_epe}. We can see that the optical flow of the car inside the red dash box becomes much better with our MotionNet. The error map in the middle column also shows that our predicted optical flow from 3D motion is much more accurate. 

\begin{table*}[!htbp]
\centering
\caption{Quantitative results of scene flow estimation and optical flow estimation on KITTI 2015 Optical Flow Train Split. Bold font indicates the best result.}
\scalebox{0.9}{
    \setlength{\tabcolsep}{8.mm}{
        % \begin{tabular}{l|cc|cc}
        \begin{tabular}{lcccc}
        \toprule  
        % \multirow{2}{*}{Methods} & \multicolumn{2}{c|}{Scene Flow} & \multicolumn{2}{c}{Optical Flow} \\
        \multirow{2}{*}{Methods} & \multicolumn{2}{c}{Scene Flow} & \multicolumn{2}{c}{Optical Flow} \\
        \cmidrule(lr){2-3}
        \cmidrule(lr){4-5}
        & D0$\downarrow$ & SF$\downarrow$ & Noc$\downarrow$ & Occ$\downarrow$ \\
        \hline
        % \hline
        GeoNet~\cite{yin2018geonet} & 49.54 & 71.32 & 8.05 & 10.81 \\
        DF-Net~\cite{zou2018df} & 46.50 & 73.30 & - & 8.98 \\
        CC~\cite{ranjan2019competitive} & 36.20 & \textbf{51.80} & - & - \\
        Monodepth2~\cite{godard2019digging} & - & - & 10.95 & 11.85 \\
        \textbf{DO3D~(ours)} & \textbf{32.58} & 54.68 & \textbf{5.49} & \textbf{7.09} \\ % stage123v3_segformer2b5_v6_pre_lr5_epoch40, weights_1
        \bottomrule
        \end{tabular}
    }
}
\label{tab:kitti_sceneflow_opticalflow}
\end{table*}

\begin{table*}[!htbp]
\centering
\caption{Ablation study on different components on the optical flow estimation task on KITTI 2015 Optical Flow Train Split. rig represents our model with object rigid motion revised, and rig+def is our full MotionNet. Bold font indicates the best result. }
\scalebox{0.9}{
    \setlength{\tabcolsep}{5.mm}{
        % \begin{tabular}{l|c|ccc|ccc}
        \begin{tabular}{lcccccc}
        \toprule
        % \multirow{2}{*}{Methods} & \multirow{2}{*}{Data} & \multicolumn{3}{c|}{Noc $\downarrow$ } &
        \multirow{2}{*}{Methods} & \multicolumn{3}{c}{Noc $\downarrow$ } &
        \multicolumn{3}{c}{Occ $\downarrow$}  \\
        \cmidrule(lr){2-4}
        \cmidrule(lr){5-7}
        % \multicolumn{1}{l|}{} & \multicolumn{1}{l|}{} & bg & fg & all & bg & fg & all \\ 
        \multicolumn{1}{l}{} & bg & fg & all & bg & fg & all \\ 
        \hline
        % \hline
        Baseline & 3.99 & 27.32 & 10.24 & 5.55 & 27.90 & 11.36 \\ % stage1_segformer2b5_v6_pre_lr5_epoch40, weights_9
        w/ rig & \textbf{3.89} & 9.96 & 5.45 & \textbf{5.49} & 10.76 & 7.11 \\ % stage12_segformer2b5_v6_pre_lr5_epoch40, weights_24
        w/ rig+def & \textbf{3.89} & \textbf{9.84} & \textbf{5.42} & \textbf{5.49} & \textbf{10.64} & \textbf{7.09} \\ % stage123v3_segformer2b5_v6_pre_lr5_epoch40, weights_8
        \bottomrule
        \end{tabular}
    }
}
\label{tab:kitti_opticalflow_epe}
\end{table*}

\begin{figure}
  \centering
	\footnotesize 
	    \centering
	    {\includegraphics[width=\linewidth]{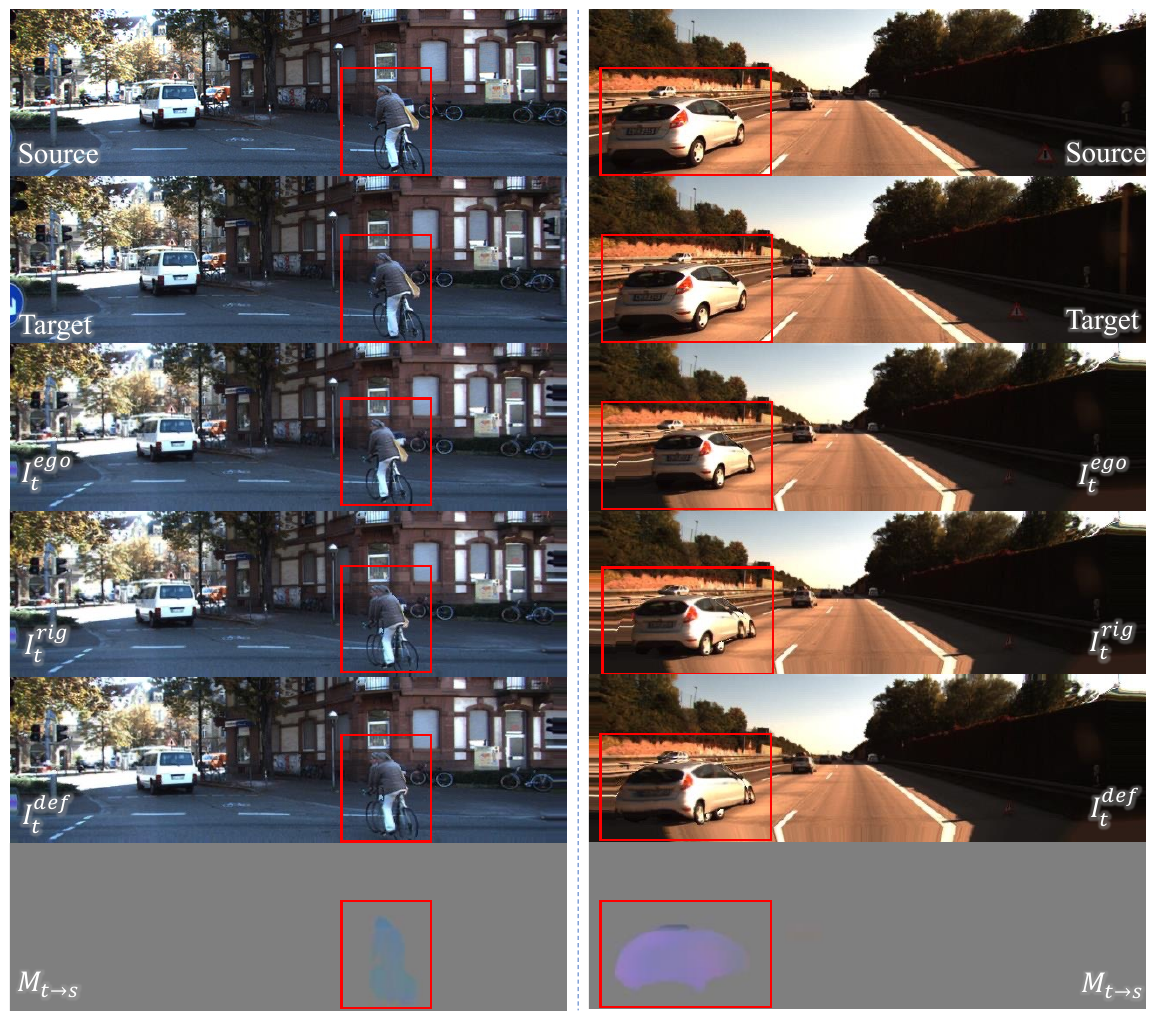}}
    \caption{Qualitative comparisons. Red boxes illustrate the major differences between different stages.}
	\label{fig:s123_comparison}
\end{figure}

\begin{figure}
    \centering
    \includegraphics[width=0.99\linewidth]{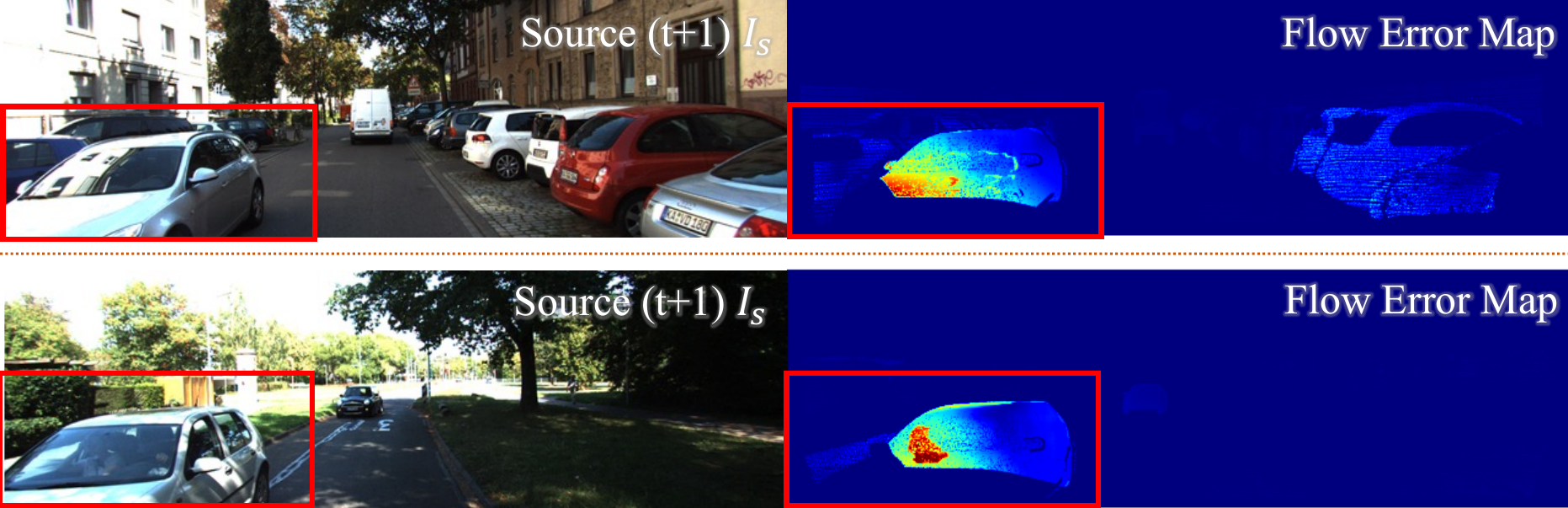}
    \caption{{Failure cases. Most of the failure cases occur in edge areas.}}
    \label{fig:failure_case}
\end{figure}

\begin{figure*}[htbp]
\centering
    \includegraphics[width=\linewidth]{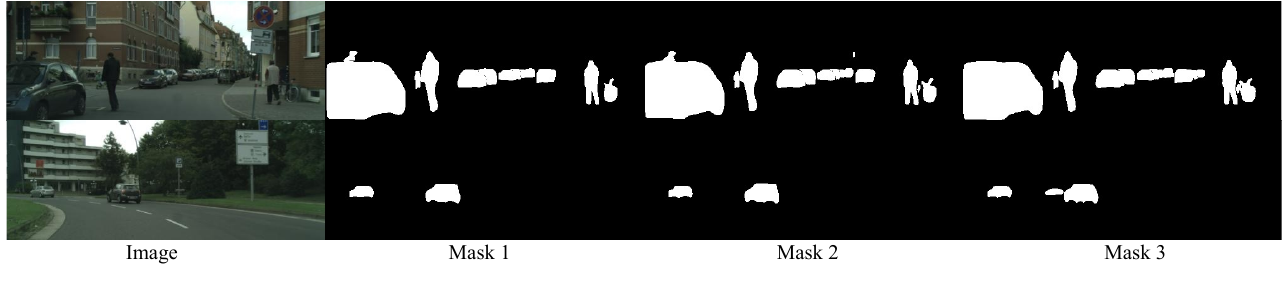}
    % \vspace{-1.0cm}
    \caption{\new{Qualitative comparisons about pre-trained Mask R-CNN models with varying accuracies on the Cityscapes dataset.}}
\label{fig:mask}
\end{figure*}

% \begin{figure*}
%   \centering
%   \subfig[]{
%     \includegraphics[width=0.93\linewidth]{pics/supp/kitti_flow_rigid_obj_gt_epe_1.pdf}
%   }
%   \subfloat[]{
%     \includegraphics[width=0.93\linewidth]{pics/supp/kitti_flow_rigid_obj_gt_epe_1.pdf}%
%   }
%   \caption{{Detailed qualitative comparisons with Monodepth2~\cite{godard2019digging}.}}
%   \label{fig:kitti_flow_epe_overall}
% \end{figure*}

\begin{figure*}[!htbp]
\centering
% \begin{subfigure}{}
%     \includegraphics[width=0.93\linewidth]{pics/supp/kitti_flow_rigid_obj_gt_epe_1.pdf}
% \label{fig:kitti_flow_epe_1}
% \end{subfigure}
\quad
\begin{subfigure}{}
    \includegraphics[width=0.93\linewidth]{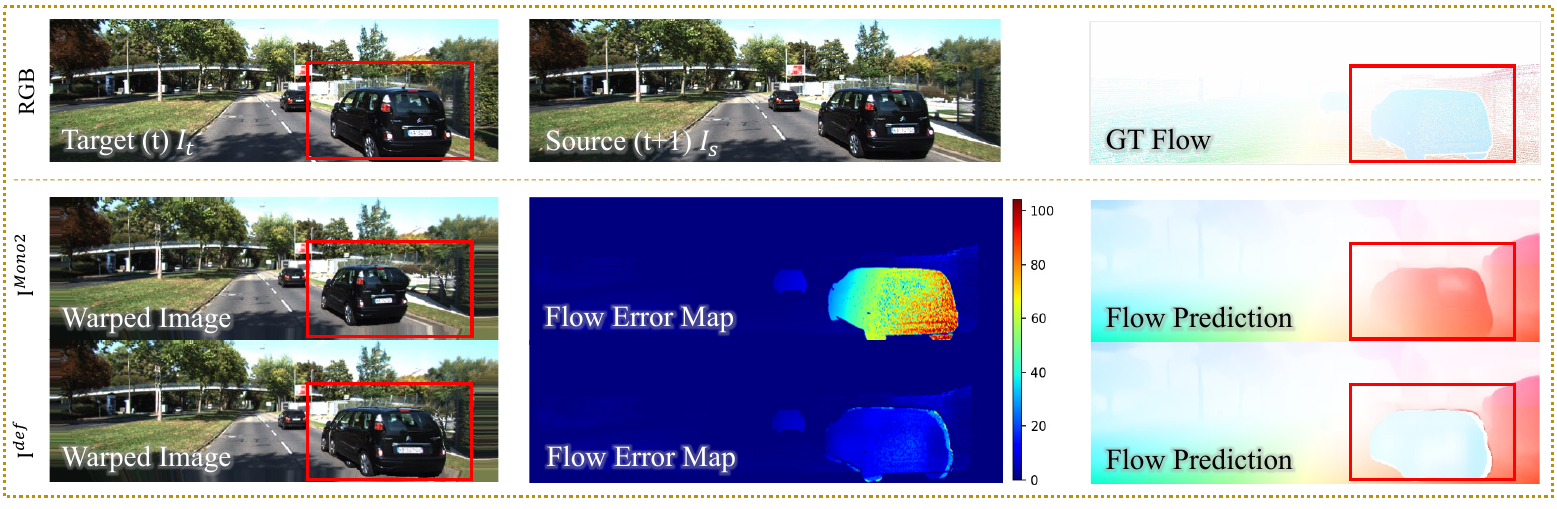}
\label{fig:kitti_flow_epe_2}
\end{subfigure}
% \begin{subfigure}{}
%     \includegraphics[width=0.93\linewidth]{pics/supp/kitti_flow_rigid_obj_gt_epe_3.pdf}
% \label{fig:kitti_flow_epe_3}
% \end{subfigure}
\quad
\begin{subfigure}{}
    \includegraphics[width=0.93\linewidth]{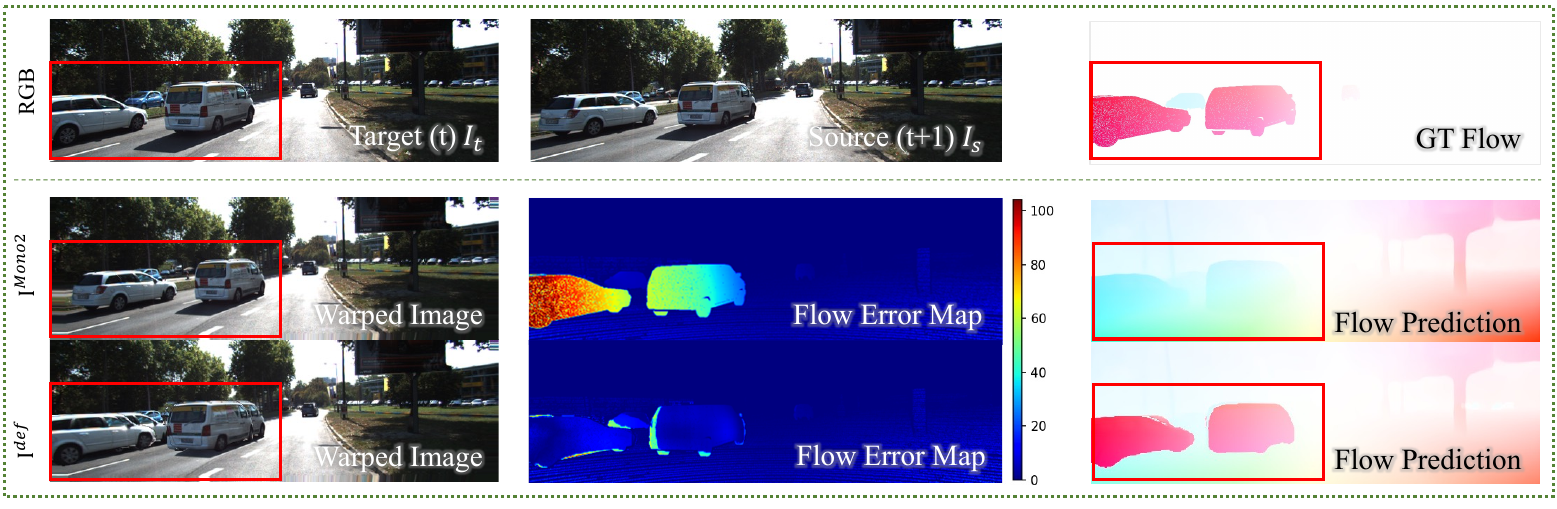}
\label{fig:kitti_flow_epe_4}
\end{subfigure}
\caption{{Detailed qualitative comparisons with Monodepth2~\cite{godard2019digging}.}}
\label{fig:kitti_flow_epe_overall}
\end{figure*}

\subsubsection{Scene Flow Evaluation}\label{sec:scene_flow_evaluation}
We also conduct scene flow experiments to further evaluate 3D motion estimation. We adopt the evaluation metric proposed in the KITTI 2015 scene flow benchmark. The metric calculates the percentage of bad pixels whose depth estimation error or optical flow estimation error exceeds the given threshold~\cite{menze2015joint, menze2015object}. 
We compare our model with several representative self-supervised monocular depth and scene flow estimation methods which are trained on monocular image sequences.
The results of official metrics are shown in Table~\ref{tab:kitti_sceneflow_opticalflow}.
It is worth mentioning that, unlike other matching-based methods that directly estimate correspondences between pixels in two frames, our optical flow and scene flow are computed based on predicted depth, camera pose, and 3D object motion. However, our model still shows superior performance in the optical flow estimation task and even outperforms self-supervised matching-based methods~\cite{ren2017unsupervised, yin2018geonet, zou2018df, ranjan2019competitive} in EPE metrics with \highlight{$5.49$} in the non-occluded region and \highlight{$7.09$} in all regions.
The reason that our model doesn't perform that well on the SF outlier metric is that our model focuses on object-wise motion modeling, while most of the SF outliers of our model lie in the background area of static scenes. The performance of static area {mainly depends on the accuracy of camera pose estimation, where a small error in pose estimation will lead to a significant error in scene flow estimation.} 
Figure~\ref{fig:3d_motion} and Figure~\ref{fig:s123_comparison} show the qualitative results of the predicted 3D motion ($\mathrm{M}_{t\rightarrow s}$): the car is moving along the z-axis with a small motion along the x-axis. 

\begin{table*}[!htbp]
\centering
\caption{{Optical flow estimation in KITTI 2015 Optical Flow Train Split. We add our model to different baseline models and train both HR-Depth and FeatDepth from scratch, labeled by $^*$. The resolutions are all 640$\times$192.}}
\scalebox{0.9}{
    \setlength{\tabcolsep}{5.mm}{
        \begin{tabular}{lcccccc}
        \toprule
        \multirow{2}{*}{Methods} & \multicolumn{3}{c}{Noc $\downarrow$ } &
        \multicolumn{3}{c}{Occ $\downarrow$}  \\
        \cmidrule(lr){2-4}
        \cmidrule(lr){5-7}
        \multicolumn{1}{l}{} & bg & fg & all & bg & fg & all\\ \hline
        % \hline
        Monodepth2~\cite{godard2019digging} & 3.59 & 29.96 & 10.42 & 5.04 & 30.72 & 11.42  \\
        \textbf{+ MotionNet} & \textbf{3.48} & \textbf{12.91} & \textbf{5.89} & \textbf{4.92} & \textbf{13.96} & \textbf{7.33} \\
        \hline 
        HR-Depth$^*$~\cite{lyu2020hr}  & 4.10 & 31.86 & 11.32 & 5.47 & 32.61 & 12.17 \\
        \textbf{+ MotionNet} & \textbf{4.03} & \textbf{11.07} & \textbf{5.77} & \textbf{5.40} & \textbf{12.16} & \textbf{7.21} \\ % weights_14
        \hline 
        %     FeatDepth~(H)~\cite{shu2020feature}  & 2.97 & 33.78 & 10.65 & 4.24 & 34.66 & 11.38 \\
        %     +$\rm{DO3D^{def}}$ & \textbf{-} & \textbf{-} & \textbf{-} & \textbf{-} & \textbf{-} & \textbf{-} \\
        % 	\hline 
        FeatDepth$^*$~\cite{shu2020feature} & \textbf{3.72} & 33.02 & 11.17 & \textbf{4.79} & 33.80 & 11.73 \\
        \textbf{+ MotionNet} & 3.76 & \textbf{16.91} & \textbf{7.00} & 4.89 & \textbf{18.04} & \textbf{8.16} \\
        \bottomrule
        \end{tabular}
    }
}
\label{tab:kitti_opticalflow_epe_baseline}
\end{table*}

\begin{table*}[!htbp]
\centering
\caption{\new{Quantitative results of monocular depth estimation on KITTI Eigen Depth Split, Cityscapes Test Split, and VKITTI2 Test Split. The comparisons are divided into twelve rows, distinguishing three different regions on three datasets.}}
\scalebox{0.71}{
    % \begin{tabular}{l|l|l|c|c|c|c|c|c|c}
    \begin{tabular}{lllccccccc}
    \toprule
    Datasets & Regions      & Methods         & Abs Rel $\downarrow$ & Sq Rel $\downarrow$ & RMSE $\downarrow$ & RMSE log $\downarrow$ & $\delta < 1.25 \uparrow$ & $\delta < 1.25^2 \uparrow$ & $\delta < 1.25^3 \uparrow$ \\
    \hline
    % \hline
    \multirow{6}*{\shortstack{KITTI Eigen Depth Split}}
    & \multirow{2}*{Foreground} 
    % & Baseline & 0.201 & 2.905 & 7.208 & 0.307 & 0.740 & 0.882 & 0.935 \\
    & Baseline & 0.157 & \textbf{1.058} & 4.814 & 0.324 & \textbf{0.823} & 0.901 & 0.929 \\
    & & \textbf{+ MotionNet} & \textbf{0.155} & 1.132 & \textbf{4.737} & \textbf{0.318} & 0.819 & \textbf{0.904} & \textbf{0.933} \\
    % & &  \rowcolor{orange!20} Improvement  & - & - & - & - & - & - & - \\
    % \cline{2-10}
    \cmidrule(lr){2-10}
    & \multirow{2}*{Background} 
    & Baseline & 0.103 & \textbf{0.749} & 4.548 & 0.167 & 0.894 & 0.969 & \textbf{0.988} \\
    & & \textbf{+ MotionNet} & \textbf{0.101} & 0.752 & \textbf{4.521} & \textbf{0.164} & \textbf{0.898} & \textbf{0.970} & \textbf{0.988} \\
    \cmidrule(lr){2-10}
    % & &  \rowcolor{orange!20} Improvement  & - & - & - & - & - & - & - \\
    % \cline{2-10} 
    & \multirow{2}*{Overall}
    & Baseline & 0.108 & \textbf{0.786} & 4.595 & 0.184 & 0.888 & \textbf{0.964} & \textbf{0.983} \\ % stage1_segformer2b5_v6_pre_lr5_epoch40, weights_8
    & & \textbf{+ MotionNet} & \textbf{0.106} & 0.797 & \textbf{4.582} & \textbf{0.181} & \textbf{0.890} & \textbf{0.964} & \textbf{0.983} \\ % stage3d_segformer2b5_v6_pre_lr5_epoch40_new, weights_1
    \hline
    % & &  \rowcolor{orange!20} Improvement  & - & - & - & - & - & - & - \\
    % \cline{1-10} 
    \multirow{6}*{\shortstack{Cityscapes Split}}
    & \multirow{2}*{Foreground} 
    % & Baseline & 0.077 & 0.239 & 1.899 & 0.099 & \textbf{0.957} & \textbf{0.993} & \textbf{0.998} \\
     & Baseline & 0.069 & 0.220 & 1.701 & 0.091 & 0.963 & \textbf{0.993} & 0.997 \\
    % & & \textbf{+ MotionNet} & \textbf{0.076} & \textbf{0.237} & \textbf{1.879} & \textbf{0.098} & 0.956 & \textbf{0.993} & \textbf{0.998} \\
     & & \textbf{+ MotionNet} & \textbf{0.068} & \textbf{0.216} & \textbf{1.704} & \textbf{0.091} & \textbf{0.962} & \textbf{0.993} & \textbf{0.998} \\
    % \cline{2-10} 
    \cmidrule(lr){2-10}
    & \multirow{2}*{Background} 
    % & Baseline & \textbf{0.124} & \textbf{1.237} & 7.043 & 0.186 & 0.838 & 0.962 & \textbf{0.989} \\
    & Baseline & 0.105 & 1.087 & 6.365 & 0.162 & 0.885 & 0.973 & \textbf{0.991} \\
    % & & \textbf{+ MotionNet}  & 0.125 & 1.263 & \textbf{6.968} & \textbf{0.185} & \textbf{0.842} & \textbf{0.963} & \textbf{0.989} \\
     & & \textbf{+ MotionNet} & \textbf{0.104} & \textbf{1.070} & \textbf{6.315} & \textbf{0.161} & \textbf{0.886} & \textbf{0.974} & \textbf{0.991} \\
    % \cline{2-10} 
    \cmidrule(lr){2-10}
    & \multirow{2}*{Overall} 
    % & Baseline & \textbf{0.121} & \textbf{1.165} & 6.838 & 0.182 & 0.846 & 0.964 & \textbf{0.990} \\
    & Baseline & 0.102 & 0.993 & 6.065 & 0.157 & 0.892 & 0.975 & \textbf{0.992} \\
    % & 0.131 & \textbf{1.244} & \textbf{7.200} & \textbf{0.195} & 0.823 & \textbf{0.957} & \textbf{0.988} \\ % stage1_segformer2b5_v6_pre_lr5_416_k2cs6w, weights_2
    & & \textbf{+ MotionNet} & 
    \textbf{0.100} & \textbf{0.977} & \textbf{6.018} & \textbf{0.156} & \textbf{0.894} & \textbf{0.976} & \textbf{0.992} \\
    \hline
    % \textbf{0.129} & 1.274 & 7.394 & \textbf{0.195} & \textbf{0.826} & 0.954 & 0.987 \\
    % \cline{1-10} 
    \multirow{6}*{\shortstack{VKITTI2 Split}}
    & \multirow{2}*{Foreground} 
    & Baseline & 0.197 & 1.967 & 7.535 & 0.255 & 0.710 & 0.922 & 0.973 \\
    & & \textbf{+ MotionNet} & \textbf{0.196} & \textbf{1.885} & \textbf{7.166} & \textbf{0.249} & \textbf{0.723} & \textbf{0.929} & \textbf{0.974} \\
    % & &  \rowcolor{orange!20} Improvement  & - & - & - & - & - & - & - \\
    % \cline{2-10} 
    \cmidrule(lr){2-10}
    & \multirow{2}*{Background} 
    & Baseline & 0.084 & 0.287 & 2.503 & 0.114 & 0.952 & 0.992 & \textbf{0.998} \\
    & & \textbf{+ MotionNet} & \textbf{0.076} & \textbf{0.253} & \textbf{2.217} & \textbf{0.103} & \textbf{0.961} & \textbf{0.996} & \textbf{0.998} \\
    % & &  \rowcolor{orange!20} Improvement  & - & - & - & - & - & - & - \\
    % % \cline{2-10} 
    \cmidrule(lr){2-10}
    & \multirow{2}*{Overall}
    & Baseline & 0.149 & 1.231 & 5.654 & 0.208 & 0.815 & 0.950 & 0.981 \\ % stage1_segformer2b5_v6_pre_lr5_lr7_k2vkclone, weigths_3
    & & \textbf{+ MotionNet} & \textbf{0.146} & \textbf{1.180} & \textbf{5.362} & \textbf{0.203} & \textbf{0.824} & \textbf{0.954} & \textbf{0.982} \\ % stage3d_segformer2b5_v6_pre_lr5_k2vkclone, weigths_1
    % & &  \rowcolor{orange!20} Improvement  & - & - & - & - & - & - & - \\
    % \cline{1-10} 
    % \hline
    % \multirow{6}*{\shortstack{\new{DAVIS Split}}}
    % & \multirow{2}*{\new{Foreground}} 
    % & \new{Baseline} & \new{0.343} & \new{\textbf{0.001}} & \new{\textbf{0.010}} & \new{0.480} & \new{0.367} & \new{0.742} & \new{0.928} \\
    % & & \new{\textbf{+ MotionNet}} & \new{\textbf{0.255}} & \new{\textbf{0.001}} & \new{\textbf{0.010}} & \new{\textbf{0.435}} & \new{\textbf{0.588}} & \new{\textbf{0.838}} & \new{\textbf{0.929}} \\
    % % \cline{2-10} 
    % \cmidrule(lr){2-10}
    % & \multirow{2}*{\new{Background}} 
    % & \new{Baseline} & \new{1.553} & \new{0.042} & \new{0.175} & \new{1.172} & \new{0.162} & \new{0.296} & \new{0.412} \\
    % & & \new{\textbf{+ MotionNet}} & \new{\textbf{0.880}} & \new{\textbf{0.030}} & \new{\textbf{0.174}} & \new{\textbf{0.904}} & \new{\textbf{0.211}} & \new{\textbf{0.396}} & \new{\textbf{0.512}} \\
    % % \cline{2-10} 
    % \cmidrule(lr){2-10}
    % & \multirow{2}*{\new{Overall}}
    % & \new{Baseline} & \new{1.400} & \new{0.037} & \new{0.171} & \new{1.159} & \new{0.160} & \new{0.286} & \new{0.403} \\
    % & & \new{\textbf{+ MotionNet}} & \new{\textbf{0.831}} & \new{\textbf{0.029}} & \new{\textbf{0.170}} & \new{\textbf{0.919}} & \new{\textbf{0.206}} & \new{\textbf{0.371}} & \new{\textbf{0.500}} \\
    \bottomrule
    \end{tabular}
    }
\label{tab:kitti_depth_fgbg}
\end{table*}

\subsubsection{Ablation Studies}\label{sec:ablation}
\noindent\textbf{Ablation on different components of MotionNet.}
To more accurately measure the contribution of each component in MotionNet, which includes both rigid motion and non-rigid deformation estimation modules, we evaluate their individual performance on two different tasks: optical flow estimation and depth estimation.

Table~\ref{tab:kitti_opticalflow_epe} presents the results of evaluations for the optical flow estimation task on the KITTI dataset.
We can conclude that: 
1) benefited from our MotionNet which estimates 3D object motion, our overall model DO3D significantly boosts the performance in foreground object regions. Compared with our baseline, our full model reduces the EPE in foreground regions by more than $60\%$, {\eg} \highlight{$9.84$} {\vs} \highlight{$27.32$} in Noc regions and \highlight{$10.64$} {\vs} \highlight{$27.90$} in Occ regions; 
2) different components of our model consistently reduce the error for foreground objects, {\eg} the original ``fg" EPE in non-occluded regions is reduced from \highlight{$27.32$} to \highlight{$9.96$} with the rigid-motion estimation component, which is further reduced to \highlight{$9.84$} with pixel-wise motion deformation component, and the trend is consistent in all regions. We also visualize the estimated 3D motion in Figure~\ref{fig:s123_comparison} ($\mathrm{M}_{t\rightarrow s}$), which demonstrates the movement of the cyclist and the car. The 3D motion map is visualized as a color image where R, G, and B correspond to motion along the $x, y$, and $z$ directions, respectively. Our intermediate results are also shown in {Figure}~\ref{fig:s123_comparison}. With the motion deformation estimator, it can be seen that the reconstructed quality for cyclists has been improved.

Table \ref{tab:kitti_depth_ablation_component} demonstrates the results of individual components concerning depth estimation on the KITTI dataset. The scarce presence of dynamic objects, encompassing both rigid and non-rigid entities, within the KITTI dataset has a minimal impact on depth estimation. 

\begin{table*}[!htbp]
\centering
\caption{\new{Component ablation study in depth estimation on the KITTI dataset. We evaluate the results within the foreground area.}}
\scalebox{0.88}{
    \new{
        \begin{tabular}{lccccccc}
        \toprule  
        Methods & Abs Rel $\downarrow$ & Sq Rel $\downarrow$ & RMSE $\downarrow$ & RMSE log $\downarrow$ & $\delta < 1.25 \uparrow$ & $\delta < 1.25^2 \uparrow$ & $\delta < 1.25^3 \uparrow$ \\ \hline 
        % \hline
        Baseline & 0.157 & 1.058 & 4.814 & 0.324 & 0.823 & 0.901 & 0.929 \\
        w/ rig  & 0.156 & 1.069 & 4.760 & 0.321 & 0.822 & 0.903 & 0.931 \\
        w/ rig+def & 0.155 & 1.132 & {4.737} & {0.318} & 0.819 & {0.904} & {0.933} \\
        \bottomrule
        \end{tabular}
    }
}
\label{tab:kitti_depth_ablation_component}
\end{table*}

% \begin{table*}[!htbp]
% % \renewcommand{\arraystretch}{1.2}
% \centering
% \caption{\new{Component ablation study in depth estimation on the DAVIS dataset \cite{perazzi2016benchmark} (Tennis sequence), denoted as D. The resolution is set to $512\times288$. We generate the pseudo-depth from DPT \cite{ranftl2021vision}.}}
% \scalebox{0.88}{
%     \new{
%         \begin{tabular}{lccccccc}
%         \toprule  
%         Methods & Abs Rel $\downarrow$ & Sq Rel $\downarrow$ & RMSE $\downarrow$ & RMSE log $\downarrow$ & $\delta < 1.25 \uparrow$ & $\delta < 1.25^2 \uparrow$ & $\delta < 1.25^3 \uparrow$ \\ \hline 
%         % \hline
%         % Baseline & 0.944 & 0.019 & 0.547 & 0.804 & 0.202 & 0.361 & 0.515 \\ 
%         Baseline & 1.400 & 0.037 & 0.171 & 1.159 & 0.160 & 0.286 & 0.403 \\
%         % w/ rig & 0.510 & 0.019 & 0.547 & 0.570 & 0.270 & 0.538 & 0.799 \\
%         w/ rig & 0.895 & 0.030 & 0.171 & 0.971 & 0.201 & 0.356 & 0.473 \\
%         % w/ rig+def & 0.484 & 0.019 & 0.547 & 0.547 & 0.288 & 0.575 & 0.835 \\
%         w/ rig+def & 0.831 & 0.029 & 0.170 & 0.919 & 0.206 & 0.371 & 0.500 \\
%         \bottomrule
%         \end{tabular}
%     }
% }
% \label{tab:davis_depth_ablation_component}
% \end{table*}

\begin{table*}[!htbp]
\centering
\caption{\new{Different architectures (CNN) in depth estimation on the KITTI dataset.}}
\scalebox{0.88}{
    \new{
        \begin{tabular}{lccccccc}
        \toprule  
        Methods & Abs Rel $\downarrow$ & Sq Rel $\downarrow$ & RMSE $\downarrow$ & RMSE log $\downarrow$ & $\delta < 1.25 \uparrow$ & $\delta < 1.25^2 \uparrow$ & $\delta < 1.25^3 \uparrow$ \\ \hline 
        % \hline
        Baseline & 0.116 & 0.873 & 4.826 & 0.194 & 0.875 & 0.959 & 0.981 \\
        \textbf{+ MotionNet} & 0.115 & 0.872 & 4.838 & 0.193 & 0.875 & 0.959 & 0.982 \\
        \bottomrule
        \end{tabular}
    }
}
\label{tab:kitti_depth_ablation_cnn}
\end{table*}

\begin{table*}[!htbp]
\centering
\caption{\new{Different architectures (CNN) in flow estimation on the KITTI dataset.}}
\scalebox{0.9}{
    \setlength{\tabcolsep}{5.mm}{
        \new{
            \begin{tabular}[t]{lcccccc}
            \toprule
            \multirow{2}{*}{Methods} & \multicolumn{3}{c}{Noc $\downarrow$ } &
            \multicolumn{3}{c}{Occ $\downarrow$}  \\  
            % \cline{2-7} 
            \cmidrule(lr){2-4}
            \cmidrule(lr){5-7}
            \multicolumn{1}{l}{} & bg & fg & all & bg & fg & all \\ \hline 
            Baseline & 3.69 & 30.91 & 10.81 & 5.06 & 31.61 & 11.69 \\
            \textbf{+ MotionNet} & 3.66 & 16.06 & 6.31 & 7.72 & 17.18 & 5.03 
            \\
            \bottomrule
            \end{tabular}
        }
    }
}
\label{tab:kitti_flow_ablation_cnn}
\end{table*}

\begin{table*}[!htbp]
\centering
\caption{\new{Ablation study on different bounding box strategies on the optical flow estimation task.}}
\scalebox{0.85}{
    \setlength{\tabcolsep}{5.mm}{
        \new{
            \begin{tabular}[t]{lccccccc}
            \toprule
            \multirow{2}{*}{Methods} & \multirow{2}{*}{Bounding box} & \multicolumn{3}{c}{Noc $\downarrow$ } &
            \multicolumn{3}{c}{Occ $\downarrow$}  \\  
            % \cline{3-8} 
            \cmidrule(lr){3-5}
            \cmidrule(lr){6-8}
            \multicolumn{1}{l}{} & \multicolumn{1}{l}{} & bg & fg & all & bg & fg & all \\ \hline 
            Baseline & - & 3.99 & 27.32 & 10.24 & 5.55 & 27.90 & 11.36 \\ 
            \textbf{+ MotionNet} & 20 pixels (current) & 3.89 & 9.84 & 5.42 & 5.49 & 10.64 & 7.09 \\ 
            \textbf{+ MotionNet} & 12.5\% & 3.90 & 10.78 & 5.63 & 5.50 & 11.71 & 7.31 \\
            \textbf{+ MotionNet} & 15.0\% & 3.88 & 10.09 & 5.42 & 5.49 & 11.02 & 7.10 \\
            \textbf{+ MotionNet} & 17.5\% & 3.90 & 9.89 & 5.44 & 5.53 & 10.81 & 7.13 \\
            \textbf{+ MotionNet} & 20.0\% & 3.92 & 10.60 & 5.49 & 5.53 & 11.82 & 7.37 \\
            \bottomrule
            \end{tabular}
        }
    }
}
\label{tab:kitti_flow_ablation_bbox_ratio}
\end{table*}

\begin{table*}[!htbp]
\centering
\caption{\new{Reproducibility study in depth estimation on the KITTI dataset.}}
\scalebox{0.79}{
    \new{
        \begin{tabular}{lccccccc}
        \toprule  
        Methods & Abs Rel $\downarrow$ & Sq Rel $\downarrow$ & RMSE $\downarrow$ & RMSE log $\downarrow$ & $\delta < 1.25 \uparrow$ & $\delta < 1.25^2 \uparrow$ & $\delta < 1.25^3 \uparrow$ \\ \hline 
        % \hline
        \textbf{DO3D (Ours)} & 0.106$\pm$0.000 & 0.798$\pm$0.010 & 4.540$\pm$0.067 & 0.181$\pm$0.001 & 0.892$\pm$0.001 & 0.965$\pm$0.001 & 0.983$\pm$0.000 \\
        \bottomrule
        \end{tabular}
    }
}
\label{tab:kitti_depth_reproduce}
\end{table*}

\begin{table*}[!htbp]
\centering
\caption{\new{Sensitivity assessment of instance segmentation network's accuracy on the KITTI dataset.}}
\scalebox{0.85}{
    \new{
        \begin{tabular}{lcccccccc}
        \toprule  
        Models & Mask AP & Abs Rel $\downarrow$ & Sq Rel $\downarrow$ & RMSE $\downarrow$ & RMSE log $\downarrow$ & $\delta < 1.25 \uparrow$ & $\delta < 1.25^2 \uparrow$ & $\delta < 1.25^3 \uparrow$ \\ \hline 
        % \hline
        Mask 1 & 32.2 & 0.106 & 0.773 & 4.497 & 0.181 & 0.893 & 0.965 & 0.983 \\
        Mask 2 & 37.2 & 0.106 & 0.797 & 4.582 & 0.181 & 0.890 & 0.964 & 0.983 \\
        Mask 3 & 39.5 & 0.105 & 0.758 & 4.493 & 0.181 & 0.893 & 0.965 & 0.983 \\
        \bottomrule
        \end{tabular}
    }
}
\label{tab:kitti_depth_ablation_mask}
\end{table*}

% \noindent\textbf{Ablation on loss functions.} 
% To evaluate the impact of different loss functions, we conduct an ablation study focusing on losses on both depth and optical flow estimation tasks. Initially, we utilize only the original photometric loss. Subsequently, we incrementally introduce additional losses to assess the performance enhancement achieved by each loss. The results of this study can be found in Table \ref{tab:kitti_depth_ablation_loss} and Table \ref{tab:kitti_flow_ablation_loss}. Remarkably, the inclusion of the proposed mask loss significantly improves the accuracy of optical flow estimation, particularly in the foreground regions. This improvement is evident in the decrease of Noc from \highlight{26.25} to \highlight{9.84} and Occ from \highlight{27.09} to \highlight{10.63}. By enabling more precise modeling of object motions, the mask loss contributes to a more accurate representation of object movements. However, the incorporation of the mask loss only brings a limited performance boost to the depth estimation task. 
% Overall, the findings from our ablation study highlight the significant impact of mask loss on optical flow estimation, while its influence on depth estimation remains relatively modest.

\noindent\textbf{Ablation on different Mask R-CNN networks.}
We have undertaken additional experiments to investigate the effectiveness of the instance segmentation network. We evaluate the robustness of our model on different Mask R-CNN networks with various precision (\ie different Mask APs on their official test sets),
which are provided in their official  \href{https://github.com/facebookresearch/detectron2/}{GitHub website}. Specifically, we utilize the model (model id: 137849600, denoted as Mask 2) with a mask AP of 37.2. Additionally, we employ two other models (model id: 137259246 and 139653917, represented as Mask 1 and Mask 3) with mask AP values of 32.2 and 39.5, respectively. Details of Mask R-CNN can be seen \href{https://github.com/facebookresearch/detectron2/blob/main/MODEL_ZOO.md}{here} and the generated segmentation results are in Figure \ref{fig:mask}. The experimental results are presented in Table \ref{tab:kitti_depth_ablation_mask}. Based on these findings, it appears that the accuracy of Mask R-CNN has a minimal influence on our model's performance. This is because our method does not rely on precise segmentation masks to be effective; instead, we utilize the bounding box surrounding the mask to identify potential dynamic regions and employ the network to learn pixel-wise non-rigid transformations which compensate for the inaccuracy of segmentation masks.

\noindent\textbf{Ablation on different architectures.}
In addition to the transformer architecture, we also conduct a comparative study using a pure CNN backbone. We integrate a CNN backbone into our DepthNet architecture, perform comprehensive evaluations including depth estimation and optical flow estimation on this network, and present the results in Table \ref{tab:kitti_depth_ablation_cnn} and Table \ref{tab:kitti_flow_ablation_cnn}. Our approach demonstrates its robustness by achieving comparable results even when transitioning from a transformer-based structure to a traditional CNN structure. These findings can show the effectiveness of our proposed method.

\noindent\textbf{Ablation on different strategies for generating bounding boxes.}
In our current approach, we extend each predicted bounding box by a fixed number of pixels (20). To explore alternative strategies for producing bounding boxes, we conduct experiments by enlarging the bounding box based on a certain ratio, ranging from 12.5\% to 20.0\%. The results, presented in Table \ref{tab:kitti_flow_ablation_bbox_ratio}, indicate that although the flexible approach aligned with the varying sizes and scales of different objects, providing a more robust and scalable solution, it results in inferior performance compared to the fixed pixel strategy. Therefore, in the final version of our model, we continue to employ the fixed pixel strategy.

\noindent\textbf{Analysis on foreground/background depth estimation.}
% The main objective of our motion decomposition module is to address the adverse impact of dynamic objects on monocular depth and motion estimation. To this end, we evaluate the performance of our proposed model against a baseline model in both foreground and background regions on four major datasets, namely KITTI Eigen Depth Split, Cityscapes Test Split, VKITTI2 Test Split, and DAVIS Test Split. For models evaluated on the Cityscapes Test split, VKITTI2 Test split, and DAVIS Test split, we finetune them on the respective Train splits based on pre-trained weights on the KITTI dataset. The results presented in Table~\ref{tab:kitti_depth_fgbg} demonstrate that our proposed model, leveraging a better representation of the underlying geometric rules, outperforms the baseline model. Notably, on the VKITTI2 and DAVIS datasets, which feature a large number of moving objects, our motion decomposition module significantly improves the depth estimation performance. On the KITTI and Cityscapes datasets, the performance gain in foreground areas is more substantial than that of the background areas, highlighting the effectiveness of our proposed model in handling dynamic objects.
The main objective of our motion decomposition module is to address the adverse impact of dynamic objects on monocular depth and motion estimation. To this end, we evaluate the performance of our proposed model against a baseline model in both foreground and background regions on three major datasets, namely KITTI Eigen Depth Split, Cityscapes Test Split, and VKITTI2 Test Split. For models evaluated on the Cityscapes Test split and VKITTI2 Test split, we finetune them on the respective Train splits based on pre-trained weights on the KITTI dataset. The results presented in Table~\ref{tab:kitti_depth_fgbg} demonstrate that our proposed model, leveraging a better representation of the underlying geometric rules, outperforms the baseline model. Notably, on the VKITTI2 dataset, which features a large number of moving objects, our motion decomposition module significantly improves the depth estimation performance. On the KITTI and Cityscapes datasets, the performance gain in foreground areas is more substantial than that of the background areas, highlighting the effectiveness of our proposed model in handling dynamic objects.

\subsubsection{More Analyses}\label{sec:analyses}
\noindent \textbf{Generalizability.}
We have conducted extensive tests to evaluate the generalizability of our method across various dimensions. Firstly, our model's performance is evaluated on three different benchmark datasets (KITTI, Cityscapes, and VKITTI2), where it consistently delivers similar performance boosts on both depth and optical flow estimation. 
% Secondly, to assess the model's generalization ability, we conduct an experiment using models trained on the Cityscapes dataset and evaluate their performance on the KITTI dataset. Results are presented in Table \ref{tab:cs2kitti_depth_generalization} and Table \ref{tab:cs2kitti_flow_generalization}, indicating that the model trained on the Cityscapes dataset performs well on the KITTI dataset. 
Besides, our developed object-wise 3D motion module is designed to be generic and compatible with various baseline architectures, demonstrating its versatility. We replace the baseline model with established depth estimation models, namely Monodepth2~\cite{godard2019digging}, HR-Depth~\cite{lyu2020hr}, and FeatDepth~\cite{shu2020feature}, and show that our model can seamlessly adapt to state-of-the-art self-supervised depth estimation models that rely on photometric loss, serving as a complementary module. To ensure a fair comparison, these models are trained from scratch, and the optical flow estimation results are presented in Table~\ref{tab:kitti_opticalflow_epe_baseline}. Notable improvements are observed in the foreground (fg) dynamic areas, thanks to the decomposed object-wise motion predictions. For instance, the fg Noc EPE is reduced from \highlight{31.86} to \highlight{11.07} when using HR-Depth+MotionNet. Similarly, we achieve a significant improvement in fg Noc/Occ for FeatDepth+MotionNet, while maintaining the original high-resolution setting, with the EPE reduced from \highlight{33.02} to \highlight{16.91}. On the other hand, the performance in the background (bg) areas is primarily influenced by the results of PoseNet and DepthNet from the baseline models, resulting in relatively modest improvement. Despite fg areas accounting for a small percentage of all pixels, the overall EPE is reduced from \highlight{11.32} to \highlight{5.77} with HR-Depth baseline Noc and from \highlight{11.17} to \highlight{7.00} with FeatDepth baseline Noc. Similar gains are observed in the Occ areas.

\noindent \textbf{Reproducibility.}
To ensure the reliability and reproducibility of our results, we have retrained our proposed model three times, and the corresponding results are presented in Table \ref{tab:kitti_depth_reproduce}. The table reveals that our model consistently achieves excellent performance across all three runs, indicating the robustness of our approach. Overall, the consistency, and robustness of our model, coupled with its impressive generalization capabilities across diverse datasets, highlight the significant potential of our proposed approach in real-world applications.

\subsubsection{More Qualitative Results}\label{sec:qualitative_results}
More qualitative results are shown in Figure~\ref{fig:kitti_flow_epe_overall},  proving that our model can handle different complex situations in motion estimation. Failure cases are shown in Figure~\ref{fig:failure_case}. In some situations, our model fails to predict the motions of objects in the edge areas. The reason is that the object-wise motion is estimated based on the target image $\mathrm{I}_t$ and the reconstructed image $\mathrm{I}_t^{ego}$. 
Due to the limitation of the backward warping mechanism in image reconstruction, the edge area of $\mathrm{I}_t^{ego}$ is often out of range and padded with the default values. Padding operation brings errors and results in bad image features, which will further have a negative impact on our object motion estimation module.

\section{Limitations} \label{sec: limitation}
Although we achieve superior performance on three tasks, we still suffer from several limitations that need future research efforts.
We note that PoseNet is prone to overfitting the autonomous driving scenario and tends to predict forward movement as most of the ego cars tend to move forward in the training dataset. This will lead to large errors in optical flow estimation which is computed based on depth and ego-motion. This problem is more severe especially when the ego-camera is static. Besides, our model still cannot fully leverage the information from neighboring frames; hence the flicking phenomenon appears when making predictions for video sequences. In future research, we would like to {improve PoseNet} and exploit temporal information to derive a more consistent model for 3D scene reconstruction.

\section{Conclusion}\label{sec:conclusion}
In this paper, we have presented a detailed analysis of the limitations of current self-supervised monocular depth estimation frameworks for dynamic scenes and revealed the causes of their poor performance in datasets containing dynamic objects. Then, we present a self-supervised framework to jointly estimate the dynamic motion of moving objects and monocular dense scene depth, which incorporates the underlying scene geometry and model into the pipeline. The core component is {the} MotionNet which combines object-wise rigid motion and pixel-wise motion deformation to represent the complicated 3D object motion. Extensive experiments on optical flow estimation, depth estimation, and scene flow estimation demonstrate the superiority of our model. We hope our analysis and encouraging results in incorporating underlying geometry processes into neural networks could inspire future investigations.

\section{Data Availability Statement}\label{sec:das}
The raw data for the KITTI dataset can be downloaded from the repository\footnote{\url{https://www.cvlibs.net/datasets/kitti/raw_data.php}}. You can use the raw dataset download script provided on their website for ease of access. For the Cityscapes dataset, downloads are accessible from their official website\footnote{\url{https://www.cityscapes-dataset.com/downloads/}}. The VKITTI2 dataset is available for download at the website\footnote{\url{https://europe.naverlabs.com/research/computer-vision/proxy-virtual-worlds-vkitti-2/}}. Visit the provided link to access the dataset and related resources.

\bibliography{sn-bibliography} % common bib file

\end{document}